\documentclass{article}

\PassOptionsToPackage{numbers, compress}{natbib}

\usepackage[eandd, preprint]{neurips_2026}

\usepackage[utf8]{inputenc}
\usepackage[T1]{fontenc}
\usepackage{hyperref}
\usepackage{url}
\usepackage{booktabs}
\usepackage{amsfonts}
\usepackage{amsmath}
\usepackage{amssymb}
\usepackage{nicefrac}
\usepackage{microtype}
\usepackage{xcolor}
\usepackage{graphicx}
\usepackage{multirow}
\usepackage[section]{placeins}
\usepackage{fancyhdr}
\usepackage{float}

\newif\ifpaperfinal
\paperfinaltrue

\title{OenoBench: A Wine-Domain Benchmark for Knowledge-Grounded \\
  Evaluation of Large Language Models}

\author{%
  Nikita Khudov, DipWSET \\
  StrategAI \\
  \texttt{nikitahudov@gmail.com}
}

\begin{document}
\makeatletter
\renewcommand{\@notice}{%
  \thispagestyle{firstpagefooter}%
}
\makeatother
\fancypagestyle{firstpagefooter}{%
  \fancyhf{}%
  \renewcommand{\headrulewidth}{0pt}%
  \renewcommand{\footrulewidth}{0pt}%
  \fancyfoot[L]{\footnotesize Preprint.}%
}
\maketitle

\begin{abstract}
We introduce \textbf{OenoBench}, a wine-domain knowledge benchmark of
\textbf{3{,}266 multiple-choice questions} across six pillars (regions,
grape varieties, viticulture, winemaking, producers, business) and four
difficulty tiers. The corpus is built from \textbf{38{,}104 atomic,
source-anchored facts} extracted by 35 provenance-verified scrapers
from government registries (INAO, TTB, OIV), peer-reviewed journals,
and Wikipedia/Wikidata. Our methodological contribution is an
\emph{LLM-driven pipeline} in which language models reformat verified
facts and audit the result, but never serve as the source of truth:
every claim traces to a URL, every question is generated by one of
five strategies across five generator families, and every question
is scored by a nine-agent audit calibrated against a human gold sheet
via Cohen's $\kappa$. Evaluating sixteen frontier configurations, we
find: (i) overall accuracy spans 53\%--84\%, led by \textbf{o3 at
83.6\%}; (ii) reasoning-mode lift concentrates in \textbf{DeepSeek R1
($+$6.8\,pp)} and is absent in Claude Opus and Gemini Pro;
(iii) Anthropic shows \textbf{$+$9\,pp self-preference} on its own
questions while Google shows \textbf{$-$8\,pp} inverse preference;
(iv) frontier open-weight models share the cost-vs-accuracy Pareto
frontier with proprietary reasoning models; and (v) every config
gains \textbf{$\approx$33\,pp} on closed-book solvable items, revealing
a parametric-recall ceiling that only the contextual slice avoids. We
release corpus, audit findings, human-review app, and construction
code under CC-BY-SA-4.0.
\end{abstract}

\section{Introduction}
\label{sec:intro}

Large language models are increasingly evaluated on knowledge-intensive
benchmarks, but the most widely-used resources --- MMLU
\citep{hendrycks2021mmlu}, GPQA \citep{rein2024gpqa}, and their
successors \citep{srivastava2023bigbench, zhong2024agieval, phan2025hle}
--- share three weaknesses: they aggregate breadth at the expense of
\emph{depth} within any single discipline; they are typically authored
by a small team using one generation pipeline, exposing systematic
blind-spots and stylistic regularities that frontier models can learn
to exploit \citep{panickssery2024selfpref, zheng2023judging}; and the
questions are increasingly suspected of contamination through web-scale
pre-training corpora \citep{sainz2023contamination,
magar2022contamination, dodge2021documenting}. As models grow more
capable, the marginal value of a benchmark depends increasingly on
\emph{how} it was constructed.

We argue that domain-specialised benchmarks built around bodies of
knowledge that are \emph{externally validated by humans} --- through
certification, regulation, or peer-reviewed practice --- offer a useful
complement to broad knowledge benchmarks. Examples already exist in
medicine \citep{jin2021medqa, jin2019pubmedqa}, law
\citep{guha2023legalbench}, and finance \citep{islam2023financebench}. We
extend this line of work to a domain that has so far received little
attention from the benchmarking community, despite being unusually
well-defined: \textbf{wine}.

\paragraph{Why wine.} Wine integrates plant biology, geology and
climate science, organic chemistry, food microbiology, sensory
neuroscience, regulatory law, and business economics. It is one of
very few domains for which inter-governmental and governmental sources
publish structured taxonomic data (INAO, OIV, TTB, regional consortia),
university research groups (UC Davis, Geisenheim, Bordeaux Sciences
Agro) produce open scholarly literature, and professional certification
bodies (WSET Levels 1--4, Court of Master Sommeliers, Institute of
Masters of Wine) define a graded competence ladder. These properties
make it possible to build a benchmark whose every question is
traceable to a verifiable source while still spanning the
recall--reasoning--application spectrum that distinguishes a strong
taster, winemaker, or buyer from a memorised glossary.

\paragraph{Contributions.}
\begin{enumerate}\itemsep -1pt
  \item \textbf{A 3{,}266-question wine benchmark} across six pillars
        and four difficulty tiers, built from 38{,}104 atomic
        provenance-tagged facts (Section~\ref{sec:dataset}).
  \item \textbf{An LLM-driven, fact-grounded generation pipeline} with
        five strategies $\times$ five generator models, per-strategy
        quotas, and an explicit closed-book pre-screen --- LLMs
        reformat verified facts, never supply them.
  \item \textbf{A nine-agent automated audit} (Section~\ref{sec:audit})
        calibrated against a human gold sheet via Cohen's $\kappa$;
        the audit removed 341 questions and re-labelled difficulty on
        1{,}259, and surfaced one agent (closed-book solvability) that
        over-reports leakage relative to humans, which we keep with
        explicit disclosure rather than drop.
  \item \textbf{A bias-aware evaluation across sixteen frontier
        configurations} (Section~\ref{sec:eval}) reporting overall
        accuracy, reasoning-mode lift, a Self-Preference Score,
        cost-efficiency Pareto, and a closed-book vs.\ source-grounded
        contrast that isolates parametric recall from contextual
        reasoning.
\end{enumerate}

\noindent OenoBench --- corpus, audit findings, construction code, and
human-review app --- ships under CC-BY-SA-4.0; release URLs,
reproducibility scripts, and the per-source licensing table are
collected in Appendix~\ref{app:artifacts}.

\section{Related Work}
\label{sec:related}

\paragraph{Broad knowledge benchmarks.} MMLU
\citep{hendrycks2021mmlu}, BIG-Bench \citep{srivastava2023bigbench},
AGIEval \citep{zhong2024agieval}, GPQA \citep{rein2024gpqa}, HELM
\citep{liang2023helm}, and Humanity's Last Exam \citep{phan2025hle}
measure factual breadth and reasoning across many subjects. We instead
maximise depth within a single externally-validated knowledge domain
and treat \emph{construction-time bias control} as a first-class
contribution rather than a downstream decontamination step.

\paragraph{Domain-specific benchmarks.} MedQA
\citep{jin2021medqa}, PubMedQA \citep{jin2019pubmedqa}, LegalBench
\citep{guha2023legalbench}, and FinanceBench
\citep{islam2023financebench} ground evaluation in regulated
professions with clear sources of truth. We adopt their
expert-syllabus / source-traceable backbone and add multi-model
generation, multi-agent audit, and self-preference reporting.
OenoBench is, to our knowledge, the first such benchmark for wine.

\paragraph{LLM-as-judge bias and self-preference.} A growing literature
documents that models tend to favour outputs stylistically similar to
their own \citep{panickssery2024selfpref, zheng2023judging,
sharma2024sycophancy}, biasing benchmarks where the same model
generates and grades. We mitigate by distributing generation across
five generator families plus deterministic templates and by reporting
per-model self-vs-other accuracy gaps as a first-class diagnostic
(Section~\ref{sec:eval-sps}).

\paragraph{Benchmark contamination and solvability.} Pre-training
exposure inflates apparent capability \citep{sainz2023contamination,
magar2022contamination, dodge2021documenting}. We mitigate by
generating from atomic facts (a fact-echo agent flags any question
whose longest common substring with the source exceeds 65\%) and by
running an explicit \emph{closed-book solvability} pre-screen and
audit (Section~\ref{sec:eval-cb}). The same audit, calibrated against
a human gold sheet, reveals that LLM judges over-report leakage by an
order of magnitude --- a finding with broader implications for the
LLM-as-judge paradigm.

\paragraph{Wine in NLP.} Wine has previously appeared in NLP via
review-text recommendation \citep{chen2014winerec}, taxonomy learning
\citep{lefever2017winenlp}, and judge-reliability studies
\citep{hodgson2017winereviews}. OenoBench is the first benchmark
targeting \emph{structured factual knowledge} of viticulture,
winemaking, and wine regions evaluated against frontier LLMs.

\section{Dataset Construction}
\label{sec:dataset}

\begin{figure}[h]
  \centering
  \includegraphics[width=0.78\linewidth]{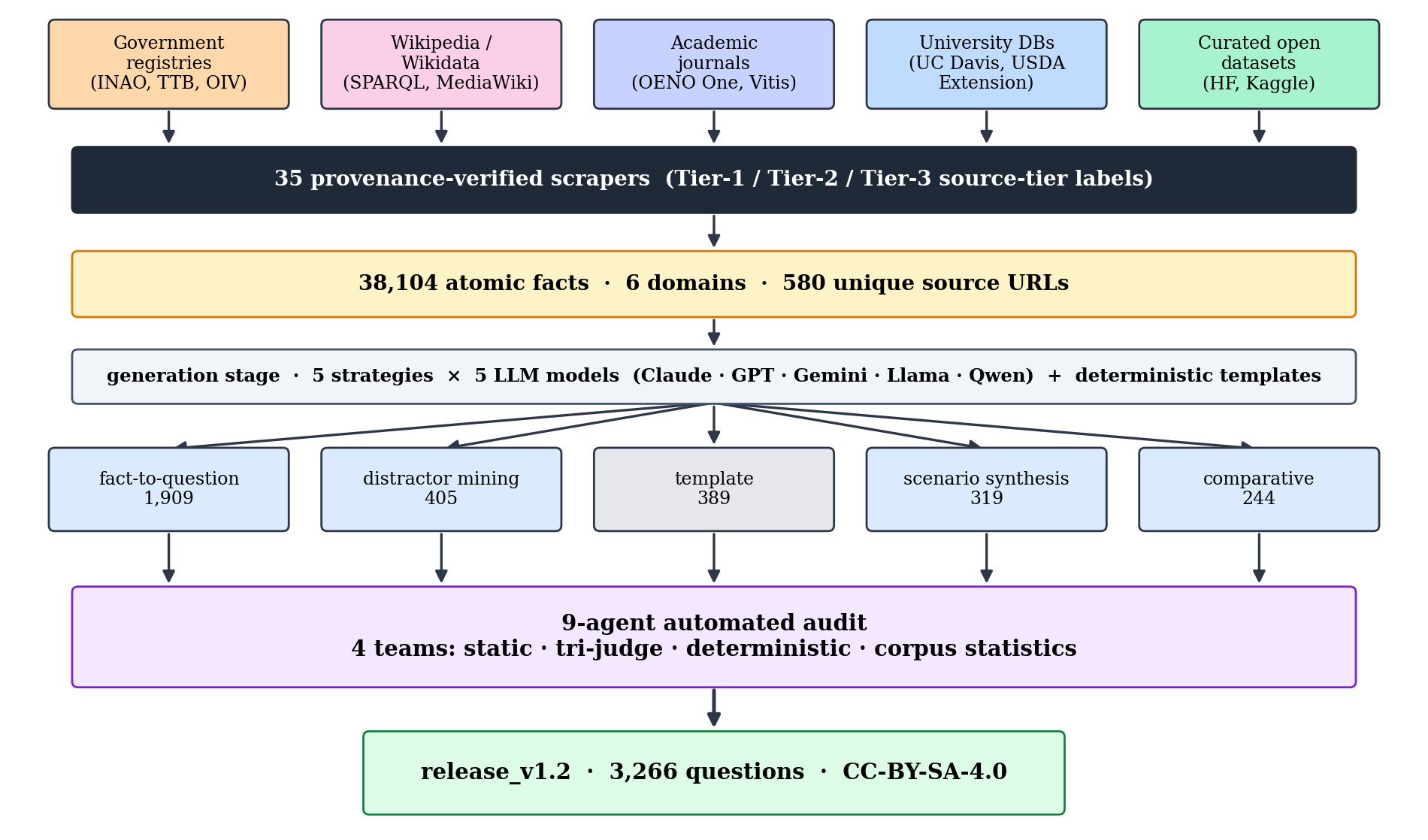}
  \caption{End-to-end OenoBench pipeline: authoritative sources $\rightarrow$
    35 scrapers $\rightarrow$ atomic facts $\rightarrow$ closed-book gate
    $\rightarrow$ nine-agent audit $\rightarrow$ release.}
  \label{fig:pipeline}
\end{figure}

OenoBench is constructed by an LLM-driven pipeline (Figure~\ref{fig:pipeline})
in two stages: (1) a \emph{fact-collection} stage that builds a corpus of
atomic, provenance-tagged statements from authoritative sources, and (2) a
\emph{question-generation} stage that converts those facts into benchmark
questions using five complementary strategies and five generator models.
The pipeline is designed around a single methodological commitment: large
language models are used as \emph{rephrasers and auditors}, never as the
source of truth. Every fact in the corpus traces to a URL with a tier-of-authority
label, every question is anchored in one or more such facts, and every
question is scored by a nine-agent automated audit (Section~\ref{sec:audit})
before release.

\subsection{Domain taxonomy}

We organise wine knowledge into six \emph{domain pillars}, chosen to align
with the WSET Level~3 and Diploma syllabi:
(i) \textbf{wine regions} (appellations, classifications, geography,
climate, soils);
(ii) \textbf{grape varieties} (ampelography, parentage, regional
distribution, sensory profile);
(iii) \textbf{viticulture} (vineyard practice, training systems, pests
and diseases, vintage effects);
(iv) \textbf{winemaking} (vinification choices, fermentation, ageing,
faults);
(v) \textbf{producers} (estates, n\'egociants, co-operatives, key
houses); and
(vi) \textbf{wine business} (markets, regulation, distribution,
economics).
Each domain has a separate target share informed by syllabus weighting
and source availability; the realised distribution is reported in
Table~\ref{tab:facts}.

\subsection{Source tiering, licensing, and non-fabrication}
\label{sec:sources}
Every fact is tagged with one of three source tiers, mirroring
evidence-based-medicine practice:
\begin{itemize}\itemsep -1pt
  \item \textbf{Tier~1 (official, 19.6\%):} government registries and
    inter-governmental bodies (INAO, OIV, TTB, regional consortia, the
    UC Davis AVA database).
  \item \textbf{Tier~2 (authoritative, 76.6\%):} Wikipedia, Wikidata
    SPARQL, wine-body and consortium websites, peer-reviewed journals
    (\emph{OENO One}, \emph{Vitis}, \emph{Australian Journal of Grape
    and Wine Research}).
  \item \textbf{Tier~3 (reliable, 3.8\%):} curated open datasets on
    HuggingFace and Kaggle, secondary reference databases.
\end{itemize}

\paragraph{Licensing and non-fabrication.}
Every source is public-record or released under a licence that permits
scraping and redistribution: Tier-1 government data under explicit
open-data licences (INAO Licence Ouverte, OIV, UC Davis AVA, TTB
public-domain), Tier-2 Wikipedia/Wikidata under CC-BY-SA / CC0 and
journals (\emph{OENO One}, \emph{Vitis}, \emph{AJEV}) open-access; the
full per-source licence table is
Table~\ref{tab:licensing} in Appendix~\ref{app:artifacts}.
A central design commitment is that \emph{no fact in the corpus comes
from an LLM's internal knowledge}: every fact is extracted by a
source-specific scraper, stamped with source URL, retrieval timestamp,
and tier label before insertion, and the fact-insertion utility
rejects records without a verified source-URL trace. We emphasise
this because LLM-generated content masquerading as scraped data is a
contamination vector that conventional dataset documentation does not
detect.

\begin{table}[h]
  \centering
  \caption{Atomic-fact corpus: 38{,}104 facts from 35 scrapers, six
    domain pillars, three source tiers.}
  \label{tab:facts}
  \small
  \begin{tabular}{lrrrr}
    \toprule
    Domain pillar & Facts & Share & Tier~1 & Tier~2/3 \\
    \midrule
    Wine regions     & 18{,}943 & 49.7\% & 23.4\% & 76.6\% \\
    Producers        &  6{,}215 & 16.3\% &  9.1\% & 90.9\% \\
    Grape varieties  &  5{,}959 & 15.6\% & 18.8\% & 81.2\% \\
    Viticulture      &  3{,}635 &  9.5\% & 12.0\% & 88.0\% \\
    Wine business    &  1{,}985 &  5.2\% & 31.4\% & 68.6\% \\
    Winemaking       &  1{,}367 &  3.6\% & 16.7\% & 83.3\% \\
    \midrule
    \textbf{Total}   & \textbf{38{,}104} & \textbf{100.0\%} &
                       \textbf{19.6\%} & \textbf{80.4\%} \\
    \bottomrule
  \end{tabular}
\end{table}

\subsection{Atomic fact extraction}
\label{sec:extraction}
Source pages are decomposed into atomic facts by a five-stage pipeline
(Figure~\ref{fig:fact_processing}, Appendix~\ref{app:fact-extraction}):
sentence splitting; reference resolution (pronouns replaced by their
entity referents); domain classification into one of the six pillars;
a length/predicate validator (5--50 words, must contain a verb, no
dangling references); and a region-keyword on-topic filter that
prevents cross-contamination (e.g.\ Austrian content in a Bordeaux
scraper). Wikidata SPARQL queries use direct country relations
(\texttt{P17}) rather than transitive parents (\texttt{P131*}), the
latter having caused severe cross-region leakage in early scrapers
(Appendix~\ref{app:methodology}).

Facts are required to be \emph{atomic} (one assertion per sentence),
\emph{entity-tagged} (subject and object linked to canonical
knowledge-graph IDs), and \emph{source-faithful} (a paraphrase, never
a verbatim copy); the last is enforced at extraction and re-checked
at audit time by agent~A3.

\subsection{Question generation: five strategies, five models}
\label{sec:generation}
A single-strategy, single-model benchmark inherits the blind-spots of
its pipeline. We mitigate this by combining \emph{five strategies} and
\emph{five generator models}, with per-strategy and per-model quotas
enforced by an orchestrator that resumes safely across runs. Strategies
are:
\begin{itemize}\itemsep -1pt
  \item \textbf{Fact-to-question (58.4\%, 1{,}909 questions):} the LLM
    rewrites a single verified fact as a multiple-choice question;
    preserves grounding but tends toward recall.
  \item \textbf{Distractor mining (12.4\%, 405):} confusable-entity
    sampling produces wrong answers that share the right answer's
    category and dimension, raising plausibility and forcing
    fine-grained discrimination.
  \item \textbf{Template (11.9\%, 389):} forty-five deterministic
    parameterised templates across the six pillars; pure entity
    substitution, zero LLM creativity --- a baseline against which
    neural generation is measured.
  \item \textbf{Scenario synthesis (9.8\%, 319):} coherent fact clusters
    are converted into applied multi-fact reasoning prompts (e.g.\ a
    sommelier service decision, a winemaker blending decision, a
    viticulturist vintage call). Domain-specific scenario types prevent
    persona-content mismatch.
  \item \textbf{Comparative (7.5\%, 244):} entity-affinity scoring pairs
    related entities (e.g.\ two Burgundy premiers crus, two Rh\^one
    grapes) and asks ``which differs in X'' or ``what do both share''.
\end{itemize}

The five generator models are Claude Opus 4.7, GPT-5, Gemini 2.5 Pro,
Llama 3.1 405B, and Qwen 3.5 235B, each accessed through a unified
OpenRouter client with consistent temperature and top-$p$. The
realised generator distribution across the released corpus is
\textbf{Qwen 20.4\%, Llama 19.3\%, Claude 19.0\%, ChatGPT 16.6\%,
Gemini 12.9\%, deterministic templates 11.9\%} (computed directly
from the 3{,}266 released questions). The orchestrator allocates
generation share dynamically based on per-pilot audit pass rates ---
generators whose questions pass audit reliably get a larger share in
subsequent rounds --- subject to a hard \textbf{per-generator cap of
21\%} to keep any single model from dominating the corpus.

\subsection{Final corpus: \texttt{release\_v1.2}}
\label{sec:release}
After generation, audit, and post-eval review, the released corpus
\texttt{release\_v1.2} contains \textbf{3{,}266 questions}. The
distribution along the domain and difficulty axes is:

\begin{table}[h]
  \centering
  \caption{Composition of the released corpus.}
  \label{tab:corpus}
  \small
  \begin{tabular}{lrrlrr}
    \toprule
    Domain & N & \% & Difficulty & N & \% \\
    \midrule
    Wine regions     & 1{,}108 & 33.9\% & L1 (entry)        & 694   & 21.2\% \\
    Grape varieties  & 766     & 23.5\% & L2 (intermediate) & 894   & 27.4\% \\
    Producers        & 515     & 15.8\% & L3 (advanced)     & 678   & 20.8\% \\
    Viticulture      & 502     & 15.4\% & L4 (expert)       & 1{,}001 & 30.6\% \\
    Wine business    & 250     &  7.7\% &                   &       &       \\
    Winemaking       & 187     &  5.7\% &                   &       &       \\
    \midrule
    \textbf{Total}   & \textbf{3{,}266} & \textbf{100\%}
                     & \textbf{Total}   & \textbf{3{,}266} & \textbf{100\%} \\
    \bottomrule
  \end{tabular}
\end{table}

\texttt{release\_v1.2} ships under CC-BY-SA-4.0 on HuggingFace with a
Croissant manifest \citep{koehn2024croissant} and a full Datasheet
\citep{gebru2021datasheets} in Appendix~\ref{app:datasheet}.

\section{Multi-Agent Quality Audit}
\label{sec:audit}

The audit pipeline runs every generated question through nine
automated agents in four teams. Each agent emits a per-question
$\{\textsc{pass}, \textsc{warn}, \textsc{fail}\}$ signal calibrated
against a human-reviewed gold sheet via Cohen's $\kappa$; agents whose
agreement falls below threshold are downweighted to advisory-only.
The released human-review web app (Appendix~\ref{app:screenshots})
collects the gold-sheet ratings, and the team-by-team architecture
(Figure~\ref{fig:audit-app}, Appendix~\ref{app:eval}) traces the flow
from candidate question to drop / relabel / keep verdict.

\subsection{Team architecture}
\label{sec:audit-arch}

Of the 14 agents (Tables~\ref{tab:agents-ab}--\ref{tab:agents-cd}),
\textbf{nine are always-run} on every candidate question; the remaining
\textbf{five are escalation-gated} and are invoked only when an upstream
agent flags a corpus-level concern. This deferred-execution policy
keeps routine audit cost bounded ($\sim$\$76 for the full corpus audit
reported here, 5h~22m wall) while preserving full coverage when a
defect class is suspected. The full audit code, prompts, and thresholds
are released alongside the corpus (Appendix~\ref{app:methodology}).

\begin{table}[H]
  \centering
  \caption{Audit agents (1/2): static and tri-judge teams.
    \textsc{always}: run on every question. \textsc{esc}:
    escalation-gated.}
  \label{tab:agents-ab}
  \scriptsize
  \setlength{\tabcolsep}{4pt}
  \renewcommand{\arraystretch}{0.95}
  \begin{tabular}{lllp{0.50\linewidth}}
    \toprule
    Team & Agent & Run & Function \\
    \midrule
    \multirow{4}{*}{A. Static}
      & A1 LexicalHygiene        & \textsc{always} & Regex sweep for vague phrasing, marketing language, meta-questions. \\
      & A2 BiasStats             & \textsc{always} & $\chi^2$ on correct-answer position; Mann--Whitney $U$ on correct-vs-distractor length. \\
      & A3 FactEcho              & \textsc{always} & Longest-common-substring ratio between question text and source fact (\textsc{fail} at LCS$\geq$0.65). \\
      & A4 TemplateFingerprint   & \textsc{always} & POS-bigram logistic regression detector for template-induced stylistic regularities (held-out AUC 0.84). \\
    \midrule
    \multirow{5}{*}{B. Tri-judge}
      & B1 TriJudgeAnswer        & \textsc{always} & Claude / GPT / Gemini panel reads question + source fact and votes on the marked-correct option. \\
      & B2 ClosedBookSolvability & \textsc{always} & Same panel, source fact \emph{withheld}: can the question be answered from world knowledge alone? \\
      & B3 UbiquityRisk          & \textsc{always} & Internationally-grown grape stem $\times$ region-class answer $\Rightarrow$ ambiguity flag. \\
      & B4 Ambiguity             & \textsc{esc}    & Tri-judge re-read flags questions with $>$1 defensible answer; invoked when B1 dissent rate elevated. \\
      & B5 VerifierSkip          & \textsc{esc}    & Self-consistency probe across re-prompted panels; invoked on disputed B1 verdicts. \\
    \bottomrule
  \end{tabular}
\end{table}

\begin{table}[h]
  \centering
  \caption{Audit agents (2/2): deterministic and corpus-statistics
    teams.}
  \label{tab:agents-cd}
  \scriptsize
  \setlength{\tabcolsep}{4pt}
  \renewcommand{\arraystretch}{0.95}
  \begin{tabular}{lllp{0.50\linewidth}}
    \toprule
    Team & Agent & Run & Function \\
    \midrule
    \multirow{4}{*}{C. Determin.}
      & C1 DistractorDifficulty  & \textsc{esc}    & Embedding-distance distractor analysis; invoked when A2 reports length skew. \\
      & C2 CategoryLeak          & \textsc{always} & Wine-type distractor validation; flags red/white/sparkling category mismatches. \\
      & C3 SourceSwap            & \textsc{esc}    & Substitutes alternate source facts to test answer stability; invoked on B1/B2 disagreement. \\
      & C4 DifficultyAudit       & \textsc{always} & Gemini Pro re-rates difficulty; \textsc{fail/warn} when delta $\geq$2 from assigned. \\
    \midrule
    \multirow{3}{*}{D. Corpus stats}
      & D1 SelfPreference        & \textsc{always} & Per-generator-family own-vs-other accuracy on a held-out probe; corpus-level signal. \\
      & D2 DedupCalibration      & \textsc{esc}    & Near-duplicate audit at cosine $\geq$0.92; invoked on suspicious cluster signals. \\
      & D3 SkewAudit             & \textsc{always} & Country / sub-domain over-representation $\chi^2$. \\
    \bottomrule
  \end{tabular}
\end{table}

\FloatBarrier
\subsection{Gold-sheet calibration}
\label{sec:gold}
A gold sheet of 50 stratified questions per release is rated independently
by three WSET-certified reviewers (Diploma, Level~3, Level~2 --- the
Diploma rater is the lead author) through the human-review web app
(Appendix~\ref{app:screenshots}); \textbf{15+ rounds} (audit pilots
v1--v16) form a calibration cycle covering eight rubrics
(answer-correctness, distractor plausibility, ambiguity,
source-faithfulness, needs-source, vague language, label correctness,
verbatim copying). For each agent we compute Cohen's $\kappa$ against
the rubric; \textbf{signals with $\kappa<0.6$ are
downweighted to advisory-only.}

The most consequential result is agent \textbf{B2~ClosedBookSolvability}:
the human solved $\approx$12\% of gold-sheet questions without the
source fact while the LLM tri-judge panel reported $\approx$83\%
($\kappa\approx0.007$). We read this as a property of the
\emph{evaluator} --- frontier LLM judges have absorbed enough wine
knowledge during pre-training to over-attribute ``world knowledge''
to non-trivial questions. We retain the 1{,}601 B2-flagged questions
with disclosure and turn the blind spot into a calibrated
memorisation-reliance diagnostic via the closed-book vs.\
source-grounded contrast (Section~\ref{sec:eval-cb}).

\subsection{Release-cycle results}
\label{sec:audit-results}
The 9-agent audit on \texttt{release\_v1.1} (3{,}670 candidate
questions) produced the verdicts summarised in
Table~\ref{tab:audit_results} (Appendix~\ref{app:audit-agents}): A1, A3,
B1, C2 and B3 between them flagged 341 distinct questions for
\textsc{drop}, while C4 re-labelled difficulty on 1{,}259 questions
and B2 flagged 1{,}601 as closed-book solvable but kept them with
disclosure.

After drops, a follow-up post-evaluation pass (Section~\ref{sec:eval}
and Appendix~\ref{app:eval}) audited the 97 questions that all 16
evaluation configs answered incorrectly: 54 were removed as defects
(wrong ground truth, all-correct options, duplicate options), 9 were
dropped on borderline-review and the remaining 34 retained as
legitimately hard. The final released corpus is 3{,}266 questions.

The C4-driven difficulty re-label was decisive: it shifted the corpus
from 14\% L3+L4 items (the original generator-assigned distribution) to
\textbf{51\% L3+L4} (Table~\ref{tab:difficulty-shift},
Appendix~\ref{app:audit-agents}), bringing the hardest-tier counts above
the entry-tier counts and substantially sharpening
difficulty-stratified analyses (Section~\ref{sec:eval-rank}).

\section{Evaluation}
\label{sec:eval}

We evaluate \textbf{16 frontier configurations} on the 3{,}266-question
release in a single end-to-end run. The slate combines six within-family
cost pairs (Claude Opus 4.7 vs.\ Claude Haiku 4.5, Gemini 2.5 Pro vs.\
Gemini 2.5 Flash, GPT-5 vs.\ GPT-5-mini, Llama 3.3 70B vs.\ Llama 3.1
8B, Qwen 2.5 72B vs.\ Qwen 2.5 7B), four reasoning-mode configurations
(o3, DeepSeek R1, Gemini 2.5 Pro thinking, Claude Opus 4.7 thinking),
and two additional standard models (DeepSeek V3, Mistral Large). Every
config sees the same 3{,}266 questions, with single-letter output
(A--D) at \texttt{max\_tokens=5} and a five-stop fallback. We organise
the evaluation around five questions, one per subsection.

\subsection{Overall capability ranking}
\label{sec:eval-rank}

\begin{figure}[h]
  \centering
  \includegraphics[width=0.72\linewidth]{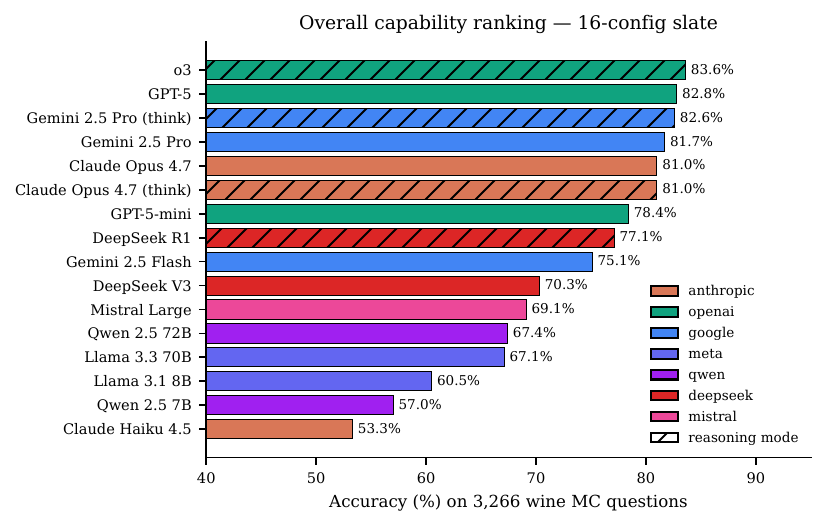}
  \caption{Overall accuracy on the 3{,}266-question corpus across 16
    configurations. Reasoning-mode configs are hatched.}
  \label{fig:leaderboard}
\end{figure}

OenoBench cleanly stratifies the slate into three bands
(Figure~\ref{fig:leaderboard}): \textbf{frontier} ($\geq$80\%, led by
\textbf{o3 at 83.6\%}, with GPT-5, Gemini 2.5 Pro, and Claude Opus 4.7
clustered within 3\,pp); \textbf{mid-tier} (mid-60s to high-70s); and
\textbf{small-tier} (low-50s to low-60s). No config performs at random;
the best-vs-worst spread is 30\,pp, comparable to GPQA
\citep{rein2024gpqa} but on a domain corpus 30$\times$ smaller and fully
provenance-tagged. Difficulty-stratified accuracies
(Appendix~\ref{app:eval}, Table~\ref{tab:full-difficulty}) confirm a
clean L1$\rightarrow$L4 gradient: 93.6\% on L1 down to \textbf{58.7\%
on L4 expert items}, with frontier configs holding 68--71\% on L4 and
small configs at 39--45\%.

\subsection{Reasoning-mode lift}
\label{sec:eval-reasoning}

Reasoning-mode lift is concentrated in a single family
(Figure~\ref{fig:reasoning}, Appendix~\ref{app:eval}). Only
\textbf{DeepSeek R1 vs.\ V3}
($+$6.8\,pp [+4.6, +8.8]) has a CI excluding zero; the three frontier
pairs (o3 vs.\ GPT-5, Gemini Pro thinking vs.\ Pro, Claude Opus
thinking vs.\ Opus) are all statistically indistinguishable from zero.
We attribute the asymmetry to headroom: V3 leaves substantial
parametric ceiling at 70.3\% which chain-of-thought can recover, while
frontier models are already near their recallable ceiling.
Per-difficulty breakdowns (Appendix~\ref{app:eval},
Table~\ref{tab:reasoning-by-diff}) show the frontier reasoning configs
gaining at most 1--2\,pp at L4 and flat elsewhere, making the
cost-vs-lift trade unfavourable except for DeepSeek
(Section~\ref{sec:eval-cost}).

\subsection{Self-preference bias}
\label{sec:eval-sps}

\begin{figure}[h]
  \centering
  \begin{minipage}[t]{0.45\linewidth}
    \includegraphics[width=\linewidth]{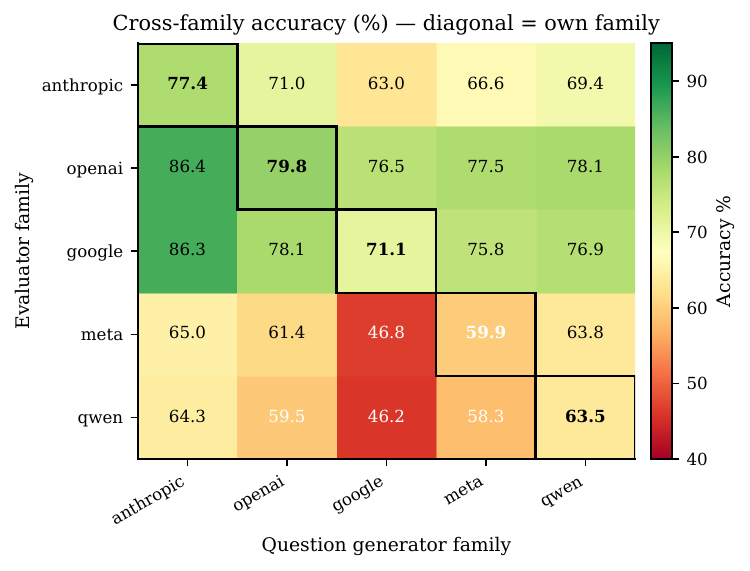}
  \end{minipage}\hfill
  \begin{minipage}[t]{0.53\linewidth}
    \includegraphics[width=\linewidth]{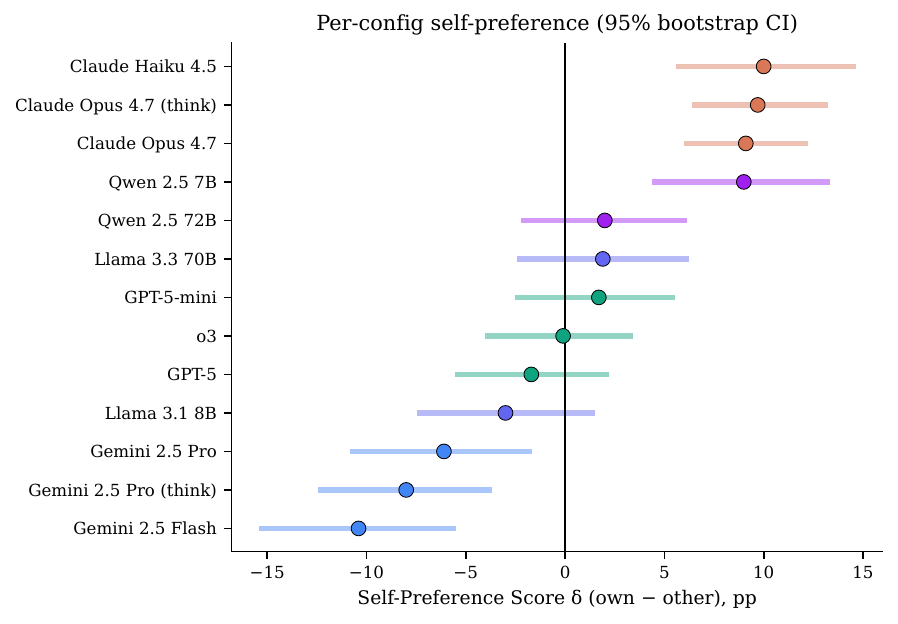}
  \end{minipage}
  \caption{Self-preference bias. \emph{Left:} cross-family accuracy
    matrix (rows = evaluator, columns = generator). \emph{Right:}
    per-config SPS $\delta = \mathrm{Acc}(\mathrm{own}) -
    \mathrm{Acc}(\mathrm{other})$ with 95\% bootstrap CI.}
  \label{fig:sps}
\end{figure}

We compute the \emph{Self-Preference Score (SPS)} per config $m$ as
$\mathrm{SPS}(m) = \mathrm{Acc}(m \mid Q_{\mathrm{own}}) -
\mathrm{Acc}(m \mid Q_{\mathrm{other}})$ where $Q_{\mathrm{own}}$
collects questions generated by $m$'s family and $Q_{\mathrm{other}}$
its complement. A pipeline that successfully neutralises generator
fingerprints should produce $|\mathrm{SPS}|$ near zero.

The result (Figure~\ref{fig:sps}) is family-dependent: \textbf{Anthropic}
clusters at $+$9 to $+$10\,pp (all CIs above zero), \textbf{OpenAI} is
statistically zero, and \textbf{Google} shows an unexpected
\emph{negative} $-$6 to $-$10\,pp cluster. The cross-family matrix
(Figure~\ref{fig:sps}, left) traces the asymmetry: Google-generated
questions are uniformly harder for everyone, while Anthropic-generated
questions are uniformly easier for Anthropic only. We interpret the
Anthropic gap as a residual stylistic fingerprint that pre-release
paraphrase passes did not neutralise, and the Google gap as
harder-by-phrasing rather than by leak; both are first-class
disclosures with the released corpus.

\subsection{Cost efficiency}
\label{sec:eval-cost}

\begin{figure}[h]
  \centering
  \includegraphics[width=0.78\linewidth]{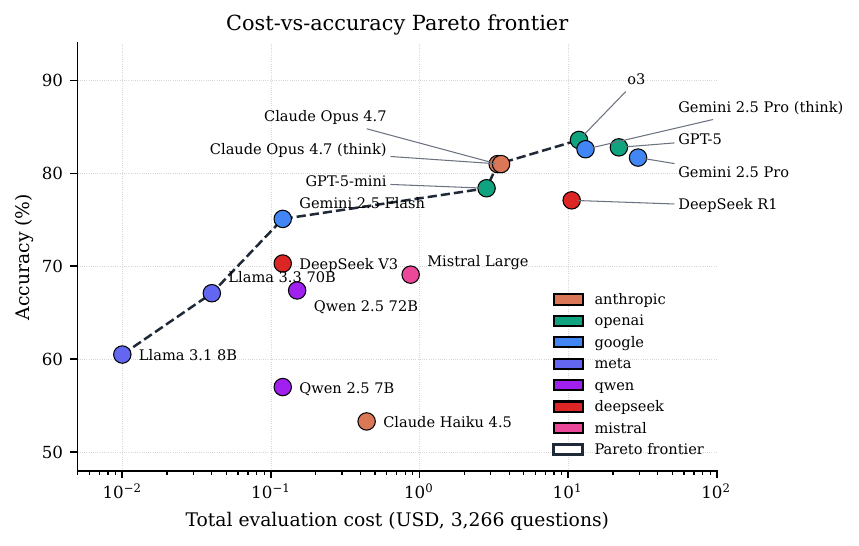}
  \caption{Total OpenRouter cost (log scale) versus accuracy on the
    full 3{,}266-question slate. Family colours match
    Figure~\ref{fig:leaderboard}; the Pareto frontier (red dashed)
    illustrates the cost-vs-quality trade-off.}
  \label{fig:pareto}
\end{figure}

The slate spans nearly four orders of magnitude in evaluation cost
(Figure~\ref{fig:pareto}). Five configs sit on the Pareto frontier
spanning the cost-quality spectrum: \textbf{Llama 3.1 8B} (60.5\,\%
/~\$0.01), \textbf{Gemini 2.5 Flash} (75.1\,\%~/~\$0.12),
\textbf{GPT-5-mini} (78.4\,\%~/~\$2.82), \textbf{Claude Opus 4.7}
(81.0\,\%~/~\$3.35), and \textbf{o3} (83.6\,\%~/~\$11.80). The two
most expensive configs (Gemini 2.5 Pro \$29.47, GPT-5 \$21.90) are
dominated by cheaper Pareto neighbours: their extra spend buys output
tokens, not better wine answers. Reasoning-mode lift
(Section~\ref{sec:eval-reasoning}) yields a Pareto win only for
DeepSeek R1, which is itself dominated by GPT-5-mini. The efficient
boundary therefore favours small proprietary plus a single reasoning
model over reasoning-everywhere or scale-everywhere strategies.

\subsection{Closed-book vs.\ source-grounded performance}
\label{sec:eval-cb}

The B2 agent reported $\sim$83\% closed-book solvability against the
human reviewer's $\sim$12\% (Section~\ref{sec:gold}) --- itself a
measurement of what frontier LLMs already \emph{know} about wine. We
exploit it by partitioning the corpus into 1{,}601 B2-flagged and
1{,}665 contextual / source-grounded questions and reporting per-config
accuracy on each slice (Figure~\ref{fig:cb}, Appendix~\ref{app:cb-figure}).

The contrast is the largest single effect we observe: the mean
per-config closed-book minus contextual gap is \textbf{$+$32.6~pp}
(range $+26.6$ to $+39.6$~pp, every CI above zero). The B2-flagged
slice overlaps heavily with parametric wine knowledge acquired in
pre-training; the contextual slice is where wine knowledge must be
reasoned through the source fact --- \textbf{Claude Opus 4.7 and o3
hold $\sim$70\% on the contextual slice, small models drop to
$\approx$45\%}, and the inter-config spread widens from 27 to 33~pp.
The contextual slice is the more discriminative test of wine reasoning;
the B2-flagged slice is a calibrated indicator of how much apparent
``wine knowledge'' is parametric recall.

\section{Applications and Future Work}
\label{sec:future}

\subsection{Closing the wine-knowledge gap in low-cost models}
Small open-weight models trail proprietary frontier ones by 20--30\,pp
on this benchmark, while the broad-benchmark gap is only 10--15\,pp
(Section~\ref{sec:eval-cost}). The 38{,}104-fact corpus is a natural
supervised-fine-tuning target via a two-stage recipe: instruction
tuning on held-out question/source-fact pairs, then LoRA adaptation
on the full fact-question-source triples in the release schema. Both
stages run on consumer GPUs without proprietary weights or
pre-training compute. The contextual slice
(Section~\ref{sec:eval-cb}) is the harder-to-game evaluation surface,
so measured gain tracks genuine wine reasoning. A
Llama-3.3-70B-class adaptation could plausibly close most of the
GPT-5-mini gap at $\approx$1\% of proprietary inference cost.

\subsection{Case-based evaluation with industry partners}
Multiple-choice tests have ceilings: a perfect OenoBench score does
not establish useful behaviour in real workflows. We are scoping an
\textbf{applied case-based dataset} with industry partners along three
personas:
\begin{itemize}\itemsep -1pt
  \item \emph{Sommelier $\rightarrow$ customer service:} pairing
    prompts rated on recommendation, justification, and substitutions.
  \item \emph{Winemaker $\rightarrow$ blending decisions:} vintage
    write-ups, lab analyses, and regulatory constraints in context;
    rated on technical correctness, compliance, and stylistic intent.
  \item \emph{Viticulturist $\rightarrow$ vineyard decisions:}
    weather, soil/disease, canopy/yield records; recommendations
    rated against decisions actually taken in the same vintage.
\end{itemize}

\subsection{Generalising the scrape-to-audit pipeline}
The pipeline modules --- tier-of-authority taxonomy, atomic-fact
extraction, multi-strategy multi-model generation with per-model
quotas, $\kappa$-calibrated multi-agent audit, and closed-book
pre-screen --- are domain-agnostic. Three near-term targets share
wine's structural features: \emph{horticulture/agronomy} (regulated
appellations, peer-reviewed journals, certification ladders);
\emph{regulated medicine} (clinical guidelines, drug formularies;
current benchmarks rely on board-exam questions misaligned with
practice); and \emph{financial regulation} (FASB, IFRS, jurisdictional
filings; CFA/CPA ladders). The released code reduces months of
expert authoring to weeks of curation.

\section{Limitations}
\label{sec:limitations}

\paragraph{Snapshot vs.\ moving target.} Wine regulation,
classifications, and producer ownership change on a multi-year cadence
(INAO appellation revisions, DOCG promotions, estate sales). The
released corpus is a snapshot dated 2026-04-01 and will require
periodic re-extraction; we release the scrapers and provenance
metadata so this is mechanically reproducible.

\paragraph{Closed-book audit calibration.} The B2 ClosedBookSolvability
agent shows $\kappa\approx0.007$ with humans on the gold sheet
(Section~\ref{sec:gold}). Rather than drop the affected questions we
keep them with explicit disclosure and report the closed-book vs.\
contextual contrast (Section~\ref{sec:eval-cb}) as a calibrated
memorisation-reliance diagnostic.

\paragraph{Self-preference is not fully neutralised.} The Anthropic
$+9$~pp positive SPS (Section~\ref{sec:eval-sps}) shows the
multi-model strategy did not fully neutralise stylistic fingerprints;
downstream Anthropic-vs-other comparisons on the full corpus should
be read with the per-config SPS in mind.

\begin{ack}
This work was independently funded by StrategAI. We thank the open
contributors to Wikipedia, Wikidata, INAO, OIV, TTB, UC Davis, USDA
Extension, and the wine consortia and academic journals
(\emph{OENO One}, \emph{Vitis}, \emph{AJEV}) whose work makes a
provenance-grounded benchmark of this scale possible. We thank the
OpenRouter team for the unified-API gateway used during construction
and evaluation.
\end{ack}

\bibliographystyle{plainnat}
\bibliography{references}

\appendix
\section{Datasheet for OenoBench}
\label{app:datasheet}

We follow the datasheet template of \citet{gebru2021datasheets}.

\subsection{Motivation}
\paragraph{For what purpose was the dataset created?}
To evaluate factual knowledge of large language models on a multi-disciplinary,
externally-validated domain (wine), and to provide a benchmark whose
construction includes explicit bias controls (multi-model generation,
multi-agent audit, self-preference scoring, closed-book vs.\ source-grounded
partition).

\paragraph{Who created the dataset?}
The dataset was created by Nikita Khudov (StrategAI) for the NeurIPS 2026
Evaluations \& Datasets Track. The lead author holds the WSET Diploma in
Wines (the highest pre-Master-of-Wine qualification of the Wine \& Spirit
Education Trust). Gold-sheet ratings used to calibrate the nine audit
agents were produced by \textbf{three WSET-certified reviewers}
(Diploma, Level~3, Level~2) including the lead author.

\paragraph{Funding.}
Independently funded by StrategAI. No external research grants were
received. Total OpenRouter API cost across the entire project ---
generation pilots, audit cycles, and the final 16-config evaluation
--- was \textbf{\$783}, paid out-of-pocket.

\subsection{Composition}
\paragraph{What do the instances represent?}
Each instance is a multiple-choice benchmark question with: question
text, 4~options with marked-correct answer, supporting fact(s) with
source URL and tier label, generator model identifier, generation
strategy, audit-agent verdicts (per-agent pass/warn/fail signals), and
(for the gold-sheet subset) human-reviewer ratings on eight rubrics.

\paragraph{How many instances are there?}
\textbf{3{,}266} questions in \texttt{release\_v1.2}. By strategy:
fact-to-question 1{,}909, distractor mining 405, template 389,
scenario 319, comparative 244. By difficulty (post-relabel):
L1 694, L2 894, L3 678, L4 1{,}001. By domain: wine regions 1{,}108,
grape varieties 766, producers 515, viticulture 502, wine business 250,
winemaking 187. The corpus is built from 38{,}104 atomic facts across
35 scrapers and 4{,}295 individually-tracked source records
spanning 56 unique top-level domains.

\paragraph{What data does each instance consist of?}
Structured Postgres records mirroring the schema documented in
Section~\ref{sec:dataset} and Appendix~\ref{app:methodology}.

\paragraph{Recommended data splits.}
The dataset ships as a single \texttt{test} split (this is an
evaluation-only benchmark). Two analytic partitions are documented and
recommended for use:
(i) the \emph{closed-book} slice (1{,}601 questions tagged
\texttt{closed\_book\_solvable}) and the \emph{contextual} slice
(1{,}665 remaining questions), partitioned by audit agent B2; and
(ii) per-generator held-out splits for self-preference analysis
(Section~\ref{sec:eval-sps}), defined dynamically by the
\texttt{generator\_family} column.

\paragraph{Errors, noise, redundancies?}
Documented in Section~\ref{sec:audit}. Per-agent fail rates and
$\kappa$ versus the human gold sheet are released alongside the
corpus. The 1{,}601 B2-flagged questions are kept with disclosure
because LLM-judge calibration with humans on closed-book solvability
is unreliable ($\kappa \approx 0.007$).

\paragraph{Self-contained, or relies on external resources?}
Self-contained: question text, supporting facts, and source URLs are
all included. Source URLs may rot over time; we mitigate by archiving
Tier-1 source pages and recording the access timestamp.

\subsection{Collection process}
\paragraph{How was the data acquired?}
By 35 provenance-verified web scrapers (released in
\texttt{src/scrapers/}). Sources span government registries (INAO,
TTB, OIV), inter-governmental bodies, university research groups
(notably UC Davis, USDA Extension), Wikipedia, Wikidata, peer-reviewed
journals (\emph{OENO One}, \emph{Vitis}, \emph{AJEV}), and curated
open datasets. Scraping was rate-limited and used the disclosed user
agent \texttt{OenoBench-Research/1.0 (academic wine benchmark)}.

\paragraph{Time frame.}
Fact extraction: 2026-03 through 2026-04. Question generation:
2026-04 through 2026-05-03. Evaluation: 2026-05-03.

\paragraph{Ethical review.}
The work is non-IRB-eligible at our institution because (a) no human
subjects were enrolled in research, (b) gold-sheet ratings were produced by
three WSET-certified reviewers including the lead author, and (c) all
source data is public-record information about commercial wine entities.
Human review of LLM-generated questions is treated as data-quality
assurance, not as human-subjects research.

\subsection{Preprocessing / cleaning / labelling}
Atomic-fact extraction with entity tagging; per-source paraphrase to
avoid verbatim copy; near-duplicate suppression at cosine~$\geq 0.92$;
nine-agent automated audit; human gold-sheet rating on a stratified
50-question subset per release, produced independently by three
WSET-certified reviewers (Diploma, Level~3, Level~2) with the lead
author resolving disagreements.

\subsection{Uses}
\paragraph{Recommended uses.} Evaluation of LLM factual knowledge in
the wine domain; ablation studies of bias-mitigation techniques in
benchmark construction; study of self-preference effects in
LLM-as-judge pipelines; calibration of automated multi-agent QA
frameworks against expert human review.

\paragraph{Discouraged uses.} Direct deployment as a wine-recommendation
or purchasing system without separate alignment, age-gating, and
jurisdictional advertising review; fine-tuning on the benchmark itself
for ranking purposes (would invalidate the benchmark for the trained
model). The scrape-to-audit pipeline should not be retargeted to
high-stakes domains (medicine, law) without comparable expert human
review of the retargeted source taxonomy.

\paragraph{Alcohol context.} Wine is an alcoholic beverage. OenoBench
is a knowledge-evaluation tool, not a consumer-facing product. A
high OenoBench score reflects factual / reasoning capability about
wine; it does not endorse downstream deployment as a
drinking-recommendation system, which would require separate
alignment for health information, age-gating, and jurisdictional
advertising rules.

\subsection{Distribution}
\paragraph{Will the dataset be distributed?}
Yes, under \textbf{CC-BY-SA-4.0} on HuggingFace at
\url{https://huggingface.co/datasets/oenobench/oenobench}. A Croissant
manifest is included \citep{koehn2024croissant}. The construction
code is on GitHub at \url{https://github.com/nikitahudov/oenobench}.

\paragraph{Citation.}
\begin{verbatim}
@inproceedings{khudov2026oenobench,
  title  = {OenoBench: A Wine-Domain Benchmark for Knowledge-Grounded
            Evaluation of Large Language Models},
  author = {Khudov, Nikita},
  booktitle = {Advances in Neural Information Processing Systems
               (NeurIPS), Datasets and Benchmarks Track},
  year   = {2026}
}
\end{verbatim}

\subsection{Maintenance}
\paragraph{Maintainer.}
Nikita Khudov, \texttt{nikitahudov@gmail.com}.

\paragraph{Update cadence.}
Wine regulation changes on a multi-year cadence; we plan annual
re-extraction. Scraper code is released so the community can
re-extract independently. Corpus versioning follows
\texttt{release\_vX.Y} (major release / minor patch) with all prior
versions retained on HuggingFace.

\paragraph{Errata.}
A public errata log will be maintained alongside the HuggingFace
release; corrections to questions, facts, audit verdicts, or
difficulty labels will be versioned and dated. Reports of factual
errors are accepted via GitHub issues and will be evaluated against
the source URL.

\section{Additional Methodology Detail}
\label{app:methodology}

This appendix expands the dataset-construction and audit pipelines:
the atomic-fact extractor (B.1), the Wikidata SPARQL choice (B.2),
question-generation prompts (B.3), the closed-book gate (B.4), and
audit architecture, agent specifications, and release-cycle results
(B.5--B.7), with the calibration philosophy that informs them (B.8).

\subsection{Atomic fact extraction pipeline}
\label{app:fact-extraction}

\begin{figure}[h]
  \centering
  \includegraphics[width=0.85\linewidth]{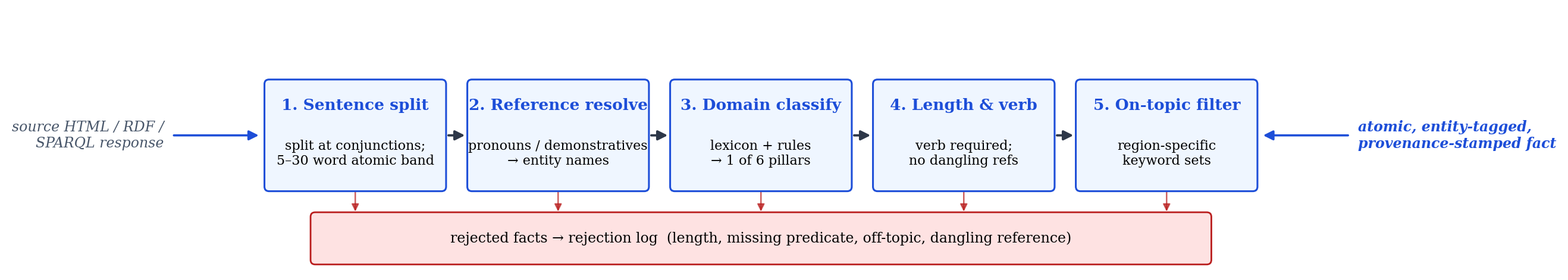}
  \caption{Atomic-fact extraction pipeline. Inputs are heterogeneous
    (HTML pages, RDF graphs, SPARQL responses, CSV dumps); outputs are
    uniform single-assertion sentences with entity tags, domain label,
    source URL and tier.}
  \label{fig:fact_processing}
\end{figure}

The atomic-fact extractor (\texttt{src/scrapers/\_fact\_processing.py})
applies five sequential filters to each candidate sentence:

\begin{enumerate}\itemsep -1pt
  \item \textbf{Sentence split.} NLTK Punkt tokeniser; further split at
    explicit conjunctions (``and'', ``but'', ``while'') when both clauses
    have an independent verb. Maximum atomic length 30~words; reject
    longer.
  \item \textbf{Reference resolution.} Pronouns and demonstratives are
    replaced with their entity referent from article context (paragraph-level
    coreference using a rule-based resolver tuned on Wikipedia
    leads). Sentences with unresolved \emph{it/this/that} are dropped.
  \item \textbf{Domain classification.} Lexicon-based + rule-based
    classifier maps each fact to one of the six domain pillars; ambiguous
    facts (no anchor term in any domain lexicon) are dropped.
  \item \textbf{Length and predicate validation.} Reject facts $<$5
    words, $>$50 words, missing a verb, or carrying dangling
    references / fragments. The 5--30 word band is the strict filter;
    the 31--50 band is kept with reduced confidence (910 facts).
  \item \textbf{On-topic filter.} Region-specific keyword sets prevent
    cross-contamination (e.g.\ Austrian content in a Bordeaux scraper).
    Fail $\Rightarrow$ drop.
\end{enumerate}

A 2026-04 audit found that 19 of an original 35 scrapers contained
hard-coded LLM-generated facts disguised as scraped data. The fix
required rebuilding all 19 against live sources; the lesson preserved
for future retargeting is that \emph{provenance auditing must run on
the scrapers themselves}, not only on the collected facts.

\subsection{Wikidata SPARQL: P17 vs P131*}
A common bug in early scrapers used the transitive administrative-parent
property (\texttt{P131*}) to bind questions to a country. This introduced
severe cross-region contamination, e.g.\ Austrian appellations appearing
in a Bordeaux scraper because the SPARQL graph traversal walked through
shared ancestor entities. Switching to the direct country relation
(\texttt{P17}) eliminated cross-country leakage at the cost of
$\sim$3\% lost coverage on borderline-region entities. The released
SPARQL templates use \texttt{P17} exclusively and we recommend the same
discipline for any pipeline retargeted to a domain with hierarchical
political subdivisions.

\subsection{Question-generation prompts}
The five strategies share a common JSON output schema enforced via
Pydantic with a 3-tier extraction fallback (markdown-fenced, raw JSON,
prefix-pattern). Per-strategy prompt summaries:

\begin{itemize}\itemsep -1pt
  \item \textbf{Fact-to-question:} "Given a single atomic wine fact,
    write a 4-option multiple-choice question whose unique correct
    answer is grounded in the fact. Paraphrase the fact; do not copy
    contiguous spans of $>$5 words. Distractors must be confusable
    same-category entities, not random alternatives."
  \item \textbf{Comparative:} given two related facts (matched on
    entity type, country/sub-domain), "ask which differs in the named
    attribute or what both share. Avoid iconic-vs-non-iconic pairings."
  \item \textbf{Scenario synthesis:} given a cluster of 3--5 facts,
    "write a brief domain-appropriate professional scenario (sommelier
    service / winemaker decision / viticulturist call /
    business/regulatory choice) with a unique correct answer that
    requires \emph{all} the cluster facts to derive."
  \item \textbf{Distractor mining:} given a fact + 3 confusable
    entities, "write a question whose 3 wrong options are these
    entities, ranked by ascending plausibility."
  \item \textbf{Template:} 45 deterministic parameter substitution
    templates; no LLM call. A subset is paraphrased post-hoc by Gemini
    Pro to defeat the A4 stylometric fingerprint detector.
\end{itemize}

Full prompt texts are in
\texttt{src/generators/\_prompts.py}.

\subsection{Closed-book gate specification}
The closed-book pre-screen (\texttt{src/generators/\_closed\_book\_gate.py})
runs a per-difficulty model:

\begin{itemize}\itemsep -1pt
  \item L1: Claude Haiku 4.5 reads the question with no source fact.
    Correct answer~$\Rightarrow$ tag \texttt{closed\_book\_solvable}.
  \item L2: Claude Sonnet 4.6 same protocol.
  \item L3: Claude Opus 4.7 same protocol; the corpus quota cap fires
    here.
  \item L4: gate is skipped by protocol; expert items are expected to
    be answerable only with the source fact.
\end{itemize}

The gate's role in generation is to enforce a corpus-level cap on
closed-book solvable items; early pilots used a 25\% cap, but it was
\textbf{raised to 50\%} after the gold-sheet calibration revealed that
the LLM panel over-attributes closed-book solvability relative to
humans (Section~\ref{sec:gold}), so the higher cap retains the items
that are in fact discriminating in evaluation. The audit's B2 agent
later re-runs the check at corpus scale on a tri-judge panel.
Failures are not dropped; they are reserved into the closed-book
slice and used for the analytic partition in
Section~\ref{sec:eval-cb}.

\subsection{Audit architecture}
\label{app:audit-architecture}

\begin{figure}[h]
  \centering
  \includegraphics[width=0.95\linewidth]{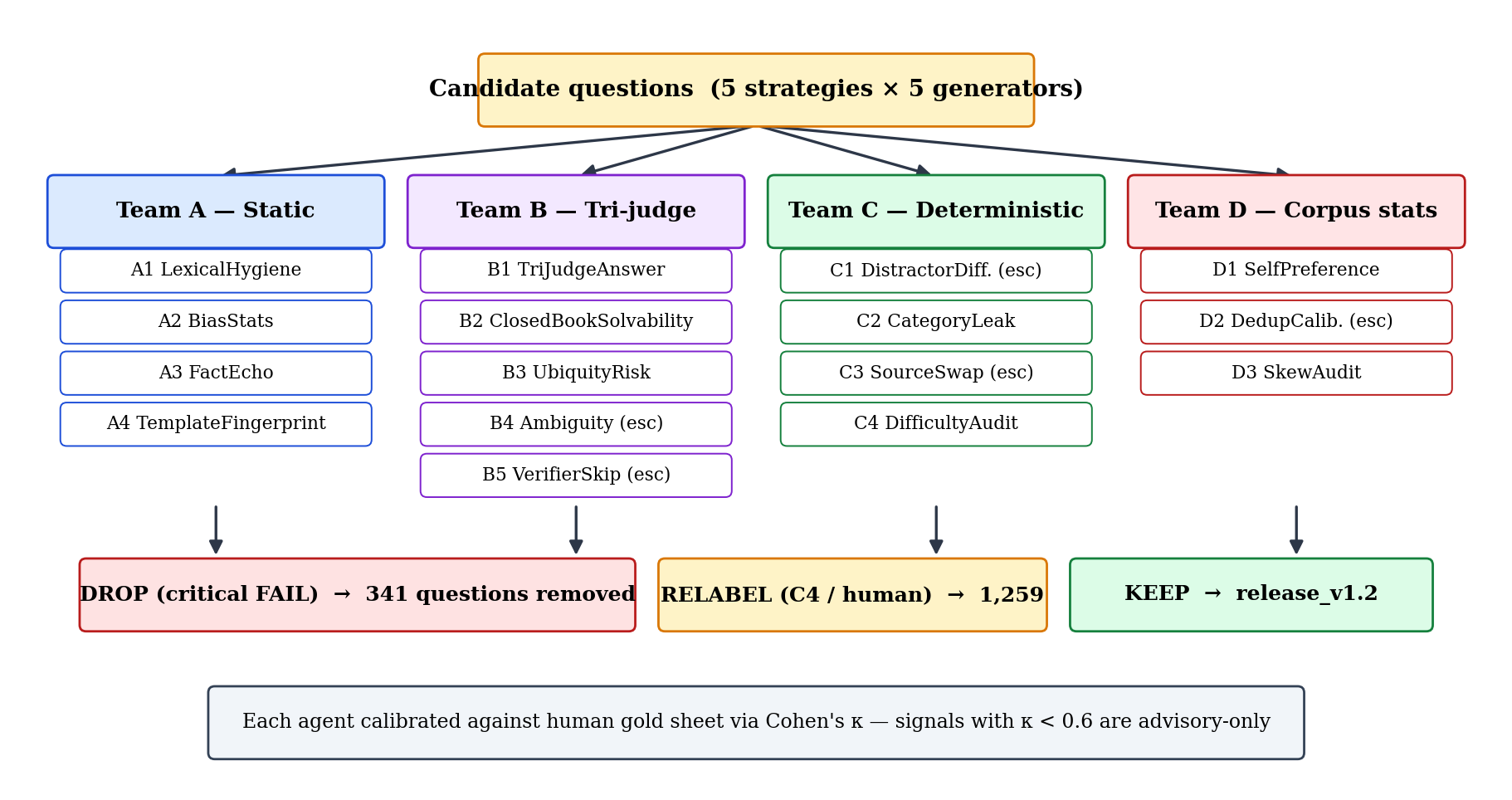}
  \caption{Multi-agent audit architecture. Candidate questions are
    fanned out to four teams; each team independently emits per-question
    verdicts. Critical \textsc{fail} signals route the question to the
    drop pool; C4 difficulty disagreements route to relabel; otherwise
    the question is kept in \texttt{release\_v1.2}. Every agent is
    calibrated against the human gold sheet via Cohen's $\kappa$.}
  \label{fig:audit-app}
\end{figure}

\subsection{Audit agent specifications}
Each agent is a versioned Python module under \texttt{src/qa/agents/}.
Agents emit
\texttt{(question\_id, agent\_id, version, severity, payload\_json)}
tuples to the \texttt{audit\_findings} table; \texttt{severity} is one
of \{\textsc{pass}, \textsc{warn}, \textsc{fail}\}. Versioning is
strict: the (\texttt{run\_id}, \texttt{question\_id}, \texttt{agent\_id},
\texttt{version}) tuple is unique, so re-running an agent at a new
version produces additive findings rather than overwriting. We document
each agent's threshold and gold-sheet $\kappa$ here.

\begin{table}[h]
  \centering
  \footnotesize
  \caption{Audit-agent thresholds and gold-sheet calibration.}
  \label{tab:agent-thresholds}
  \begin{tabular}{lllr}
    \toprule
    Agent & Signal & Threshold & $\kappa$ vs.\ gold \\
    \midrule
    A1 LexicalHygiene      & vague/marketing regex hits & $\geq$1 hit $\Rightarrow$ \textsc{warn}; $\geq$3 $\Rightarrow$ \textsc{fail} & 0.71 \\
    A2 BiasStats           & $\chi^2$ position; M--W length & $p<0.01 \Rightarrow$ \textsc{fail} & 0.65 \\
    A3 FactEcho            & LCS / fact length          & $\geq$0.45 $\Rightarrow$ \textsc{warn}; $\geq$0.65 $\Rightarrow$ \textsc{fail} & 0.83 \\
    A4 TemplateFingerprint & POS-bigram logreg held-out AUC & $>0.84$ corpus-level $\Rightarrow$ \textsc{warn} & 0.62 \\
    B1 TriJudgeAnswer      & majority disagrees with key & 2/3 disagree $\Rightarrow$ \textsc{fail} & 0.78 \\
    B2 ClosedBookSolv.     & majority solves no-source & 2/3 solve $\Rightarrow$ \textsc{warn}; 3/3 $\Rightarrow$ \textsc{fail} & \textbf{0.007} \\
    C2 CategoryLeak        & wine-type mismatch in distractors & any $\Rightarrow$ \textsc{fail} & 0.91 \\
    C4 DifficultyAudit     & $|\Delta|$ assigned vs.\ rated & $\geq$1 \textsc{warn}; $\geq$2 \textsc{fail} & 0.74 \\
    B3 UbiquityRisk        & ubiquitous-grape stem $\times$ region answer & both $\Rightarrow$ \textsc{warn} & 0.69 \\
    \bottomrule
  \end{tabular}
\end{table}

The B2 row is the calibration finding discussed in
Section~\ref{sec:gold}: the LLM panel and the human reviewer disagree
on what counts as ``world-knowledge solvable'', because frontier LLMs
have absorbed substantial wine knowledge during pre-training. We retain
the agent as a measurement of the LLM panel's prior, and report the
contextual / closed-book partition (Section~\ref{sec:eval-cb}) that
turns the disagreement into a useful diagnostic.

\subsection{Release-cycle results}
\label{app:audit-agents}

Table~\ref{tab:audit_results} summarises the verdicts produced when
the 9-agent audit was applied to \texttt{release\_v1.1} (3{,}670
candidate questions); these are the numbers cited in
Section~\ref{sec:audit-results}.

\begin{table}[h]
  \centering
  \caption{Audit findings on \texttt{release\_v1.1} and the resulting
    drop / re-label policy.}
  \label{tab:audit_results}
  \footnotesize
  \begin{tabular}{lrl}
    \toprule
    Audit signal & Triggered FAIL & Action \\
    \midrule
    A1 LexicalHygiene (vague phrasing)               & 60   & drop \\
    A3 FactEcho (verbatim copy)                      & 63   & drop \\
    B1 TriJudgeAnswer (wrong answer key)             & 47   & drop \\
    C2 CategoryLeak (wine-category leak)             & 9    & drop \\
    B3 UbiquityRisk (grape $\times$ region ambiguity)& 183  & drop \\
    \midrule
    Distinct dropped (with overlap)                  & \textbf{341} & \\
    \midrule
    C4 DifficultyAudit (relabel from 1{,}252 LLM hits + 7 human)
                                                     & 1{,}259 & re-label \\
    B2 ClosedBookSolvability (kept w/ disclosure)     & 1{,}601 & disclose \\
    \bottomrule
  \end{tabular}
\end{table}

Table~\ref{tab:difficulty-shift} shows the corpus-level effect of the
C4 difficulty re-label, which shifted the corpus from 14\% L3+L4 items
to 51\% L3+L4.

\begin{table}[h]
  \centering
  \caption{Difficulty distribution before and after C4 re-label.}
  \label{tab:difficulty-shift}
  \footnotesize
  \begin{tabular}{lrrr}
    \toprule
    & Generator-assigned & Post-relabel & $\Delta$ \\
    \midrule
    L1 (entry)        & 1{,}261 (38\%) & 694   (21\%) & $-567$ \\
    L2 (intermediate) & 1{,}559 (47\%) & 894   (27\%) & $-665$ \\
    L3 (advanced)     &   218 ( 7\%) & 678   (21\%) & $+460$ \\
    L4 (expert)       &   291 ( 9\%) & 1{,}001 (31\%) & $+710$ \\
    \bottomrule
  \end{tabular}
\end{table}

\subsection{Calibration philosophy}
\label{app:calibration-philosophy}
We emphasise the design choice underlying these numbers. A naive
``automated audit'' interpretation would have dropped all 1{,}601
B2-flagged questions, which would have removed nearly half the corpus
and biased the remainder toward whatever idiosyncratic items the LLM
panel happened not to know. By calibrating B2 against humans and
treating it as an evaluator-side signal, we preserve dataset value while
giving downstream users a principled way to interpret leakage. The
audit's role, on this reading, is not to be the final arbiter of
quality, but to surface signals \emph{that humans validate} as
quality-relevant, and to disclose clearly the signals where humans and
LLMs disagree.

\section{Full Evaluation Tables}
\label{app:eval}

\raggedbottom
\setlength{\textfloatsep}{8pt plus 2pt minus 2pt}
\setlength{\intextsep}{8pt plus 2pt minus 2pt}
\setlength{\floatsep}{8pt plus 2pt minus 2pt}
\setlength{\abovecaptionskip}{4pt}
\setlength{\belowcaptionskip}{2pt}

This appendix gives the per-config tables and supporting figures
behind each Section~\ref{sec:eval} subsection: reasoning-mode lift
(C.1), self-preference bootstrap CIs (C.2), the closed-book vs.\
source-grounded contrast (C.3), and per-domain (C.4), per-strategy
(C.5), and per-difficulty (C.6) breakdowns of the overall ranking.
The cost ledger that supports Section~\ref{sec:eval-cost} is in
Appendix~\ref{app:artifacts}.

\subsection{Reasoning-mode lift}
\label{app:reasoning}

\begin{figure}[H]
  \centering
  \includegraphics[width=0.7\linewidth]{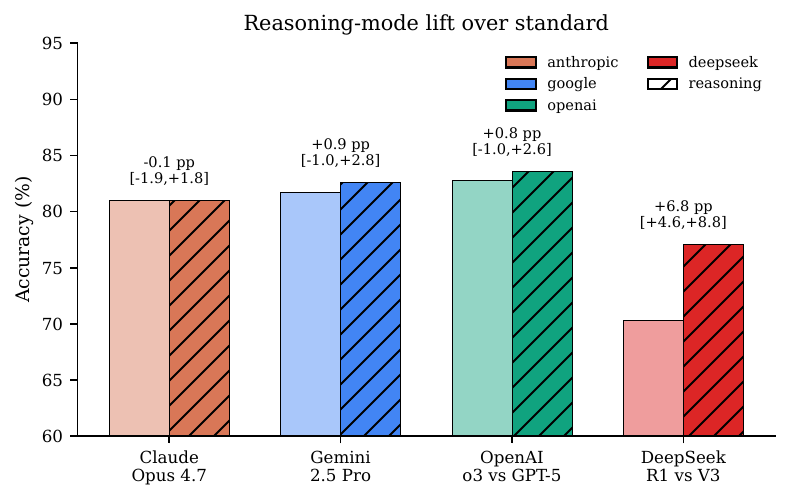}
  \caption{Reasoning-mode vs.\ standard accuracy for four families;
    $\delta$ in pp with 95\% bootstrap CI (1{,}000 resamples). Only
    DeepSeek's CI excludes zero.}
  \label{fig:reasoning}
\end{figure}

\begin{table}[H]
  \centering
  \footnotesize
  \caption{Reasoning lift per difficulty tier ($\delta$ = thinking $-$ standard, pp).}
  \label{tab:reasoning-by-diff}
  \begin{tabular}{lrrrr}
    \toprule
    Pair & L1 & L2 & L3 & L4 \\
    \midrule
    Claude Opus 4.7 (think vs std)        & $+$0.3 & $-$0.2 & $+$0.3 & $-$0.5 \\
    Gemini 2.5 Pro (think vs std)         & $+$0.5 & $+$1.0 & $+$1.0 & $+$0.9 \\
    OpenAI o3 vs GPT-5                    & $+$0.1 & $-$1.0 & $+$1.9 & $+$2.0 \\
    DeepSeek R1 vs V3                     & $+$1.7 & $+$8.3 & $+$5.7 & $+$9.7 \\
    \bottomrule
  \end{tabular}
\end{table}

\FloatBarrier
\subsection{Self-preference bootstrap CIs}

\begin{table}[H]
  \centering
  \footnotesize
  \caption{Per-config Self-Preference Score with 95\% bootstrap CI
    (1{,}000 resamples). $n$ columns are the question counts in each
    pool: own~$=$~questions whose generator family matches the
    config's family; other~$=$~questions whose generator is a
    different non-template LLM. Templates ($n{=}389$) are excluded
    from both pools, so own~$+$~other~$=$~2{,}877 across every row.
    DeepSeek and Mistral have no own-family questions in
    \texttt{release\_v1.2} and are omitted.}
  \label{tab:sps-full}
  \begin{tabular}{lrrrrrr}
    \toprule
    Config & own $n$ & other $n$ & own (\%) & other (\%) & $\delta$ (pp) & 95\% CI \\
    \midrule
    claude-haiku-4.5             & 619 & 2{,}258 & 58.8 & 48.8 & $+10.0$ & [$+5.7$, $+14.5$] \\
    claude-opus-4.7              & 619 & 2{,}258 & 86.6 & 77.5 & $+9.1$  & [$+6.1$, $+12.1$] \\
    claude-opus-4.7-thinking     & 619 & 2{,}258 & 86.9 & 77.2 & $+9.7$  & [$+6.5$, $+13.1$] \\
    qwen-2.5-7b                  & 667 & 2{,}210 & 60.9 & 51.9 & $+9.0$  & [$+4.5$, $+13.2$] \\
    qwen-2.5-72b                 & 667 & 2{,}210 & 66.1 & 64.1 & $+2.0$  & [$-2.1$, $+6.0$] \\
    llama-3.3-70b                & 629 & 2{,}248 & 65.5 & 63.6 & $+1.9$  & [$-2.3$, $+6.1$] \\
    gpt-5-mini                   & 542 & 2{,}335 & 77.9 & 76.1 & $+1.7$  & [$-2.4$, $+5.4$] \\
    o3                           & 542 & 2{,}335 & 81.9 & 82.1 & $-0.1$  & [$-3.9$, $+3.3$] \\
    gpt-5                        & 542 & 2{,}335 & 79.7 & 81.4 & $-1.7$  & [$-5.4$, $+2.1$] \\
    llama-3.1-8b                 & 629 & 2{,}248 & 54.2 & 57.2 & $-3.0$  & [$-7.3$, $+1.4$] \\
    gemini-2.5-pro               & 420 & 2{,}457 & 75.0 & 81.1 & $-6.1$  & [$-10.7$, $-1.8$] \\
    gemini-2.5-pro-thinking      & 420 & 2{,}457 & 74.0 & 82.0 & $-8.0$  & [$-12.3$, $-3.8$] \\
    gemini-2.5-flash             & 420 & 2{,}457 & 64.3 & 74.7 & $-10.4$ & [$-15.3$, $-5.6$] \\
    \bottomrule
  \end{tabular}
\end{table}

\begin{table}[H]
  \centering
  \footnotesize
  \caption{Per-tier Self-Preference Score $\delta$ (pp). The overall
    SPS in Table~\ref{tab:sps-full} is decomposed by the question's
    post-relabel difficulty tier; each cell is own-Acc minus other-Acc
    on the questions of that tier. \textbf{Bold} marks cells whose
    95\% bootstrap CI excludes zero.}
  \label{tab:sps-by-tier}
  \begin{tabular}{llrrrr}
    \toprule
    Config & Family & L1 $\delta$ & L2 $\delta$ & L3 $\delta$ & L4 $\delta$ \\
    \midrule
    claude-haiku-4.5         & anthropic & $+3.1$           & $\mathbf{+10.2}$ & $+2.0$           & $+8.2$            \\
    claude-opus-4.7          & anthropic & $+0.2$           & $\mathbf{+8.5}$  & $+4.1$           & $\mathbf{+9.5}$   \\
    claude-opus-4.7-thinking & anthropic & $-0.6$           & $\mathbf{+10.2}$ & $+3.4$           & $\mathbf{+10.9}$  \\
    qwen-2.5-7b              & qwen      & $\mathbf{+12.3}$ & $+6.6$           & $+7.2$           & $\mathbf{+10.4}$  \\
    qwen-2.5-72b             & qwen      & $+2.4$           & $-4.3$           & $+0.0$           & $\mathbf{+7.7}$   \\
    llama-3.3-70b            & meta      & $-0.8$           & $-3.8$           & $-5.5$           & $+6.8$            \\
    gpt-5-mini               & openai    & $\mathbf{+3.1}$  & $\mathbf{+10.7}$ & $+4.4$           & $-0.7$            \\
    o3                       & openai    & $\mathbf{+2.6}$  & $\mathbf{+6.9}$  & $+4.0$           & $-3.2$            \\
    gpt-5                    & openai    & $\mathbf{+2.9}$  & $\mathbf{+7.2}$  & $+0.8$           & $-6.9$            \\
    llama-3.1-8b             & meta      & $-1.3$           & $\mathbf{-14.6}$ & $-6.9$           & $+1.9$            \\
    gemini-2.5-pro           & google    & $-1.0$           & $+4.5$           & $+0.7$           & $\mathbf{-10.3}$  \\
    gemini-2.5-pro-thinking  & google    & $\mathbf{+1.3}$  & $-0.3$           & $+5.8$           & $\mathbf{-14.1}$  \\
    gemini-2.5-flash         & google    & $-0.2$           & $-6.3$           & $-0.8$           & $\mathbf{-10.6}$  \\
    \bottomrule
  \end{tabular}
\end{table}

\paragraph{Per-tier reading.} The Anthropic positive SPS is
\emph{not} concentrated at any single difficulty tier — it spikes at
both L2 ($+8.5$ to $+10.2$ across the three Claude configs, all CIs
above zero) and L4 ($+9.5$ to $+10.9$, two of three CIs above zero),
while L1 (near ceiling) and L3 are flat. A pure question-difficulty
asymmetry would predict the gap to widen monotonically as accuracy
falls; instead the gap appears at \emph{both} ends of the non-trivial
range, which is more consistent with a residual stylistic fingerprint
than with Anthropic having generated easier questions. Google's
negative SPS is overwhelmingly an L4 phenomenon (Gemini Pro / Pro
thinking / Flash all $-10$ to $-14$~pp at L4, with CIs excluding zero;
L1--L3 within $\pm 6$~pp), consistent with Gemini-authored hardest-tier
questions being uniformly hard rather than the model being penalised
on its own style. OpenAI's near-zero overall SPS is the result of
positive L1/L2 and negative L4 cells averaging out, not stylistic
neutrality.

\FloatBarrier
\subsection{Closed-book vs.\ source-grounded performance}
\label{app:cb-figure}

\begin{figure}[H]
  \centering
  \includegraphics[width=0.7\linewidth]{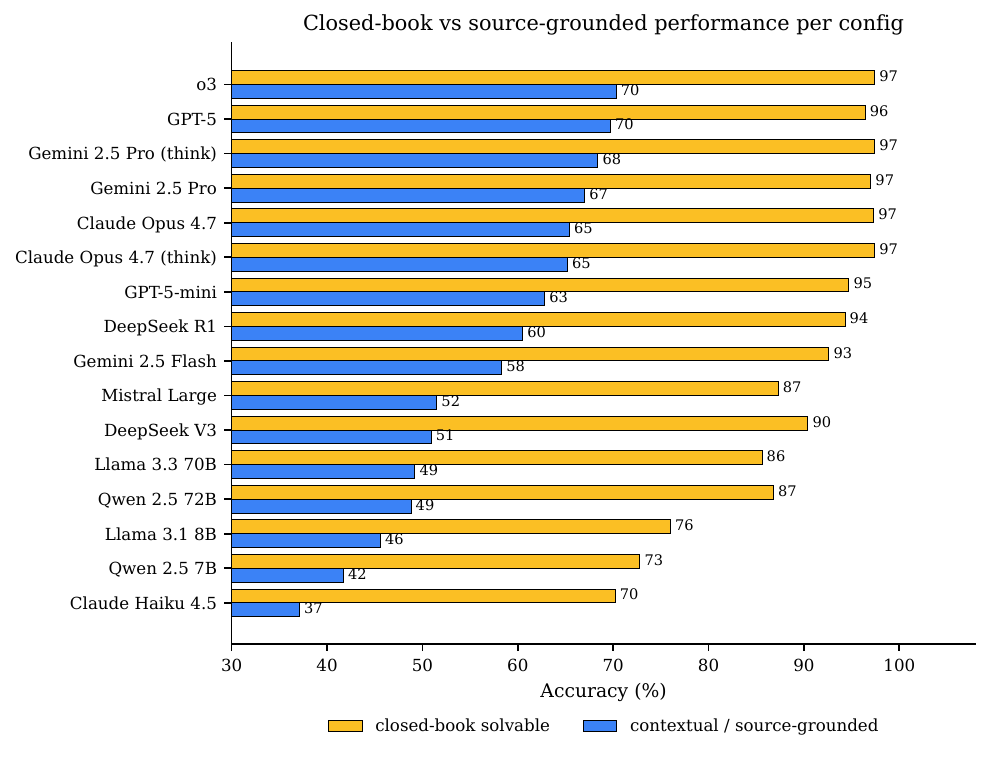}
  \caption{Per-config accuracy on B2-flagged closed-book solvable
    questions ($n{=}1{,}601$) vs.\ contextual / source-grounded
    questions ($n{=}1{,}665$); every config gains $+26.6$ to
    $+39.6$~pp on the closed-book slice.}
  \label{fig:cb}
\end{figure}

\begin{table}[H]
  \centering
  \footnotesize
  \caption{Closed-book vs.\ contextual accuracy gap (pp) within each
    difficulty tier. A positive gap means the model scores higher on
    B2-flagged closed-book solvable items than on contextual items at
    the same tier. Per-tier B2-flagged / contextual question counts:
    L1 680 / 13, L2 333 / 561, L3 472 / 206, L4 116 / 885. The Mean
    row reports 95\% bootstrap CIs (1{,}000 resamples;
    question-level resampling within each pool).}
  \label{tab:cb-by-tier}
  \begin{tabular}{lrrrr}
    \toprule
    Config & L1 gap & L2 gap & L3 gap & L4 gap \\
    \midrule
    claude-opus-4.7              & $+5.8$  & $+30.0$ & $+35.3$ & $+31.0$ \\
    claude-haiku-4.5             & $+13.0$ & $+33.5$ & $+22.6$ & $+25.4$ \\
    gpt-5                        & $-1.8$  & $+23.2$ & $+21.0$ & $+28.0$ \\
    gpt-5-mini                   & $-2.1$  & $+32.8$ & $+24.0$ & $+26.1$ \\
    gemini-2.5-pro               & $-1.5$  & $+28.4$ & $+27.8$ & $+27.8$ \\
    gemini-2.5-flash             & $+11.0$ & $+33.9$ & $+27.4$ & $+28.5$ \\
    llama-3.3-70b                & $+7.4$  & $+37.4$ & $+23.7$ & $+19.1$ \\
    llama-3.1-8b                 & $-15.4$ & $+29.0$ & $+17.2$ & $+11.6$ \\
    deepseek-v3                  & $+3.6$  & $+41.0$ & $+26.6$ & $+31.4$ \\
    qwen-2.5-72b                 & $-6.9$  & $+38.0$ & $+25.0$ & $+31.5$ \\
    qwen-2.5-7b                  & $-2.7$  & $+30.9$ & $+14.4$ & $+21.3$ \\
    mistral-large-2411           & $+2.1$  & $+34.4$ & $+28.6$ & $+23.7$ \\
    o3                           & $-1.6$  & $+24.8$ & $+22.4$ & $+29.6$ \\
    gemini-2.5-pro-thinking      & $-0.9$  & $+27.2$ & $+25.1$ & $+26.8$ \\
    deepseek-r1                  & $-2.5$  & $+33.5$ & $+28.5$ & $+28.2$ \\
    claude-opus-4.7-thinking     & $+13.9$ & $+30.7$ & $+32.3$ & $+32.6$ \\
    \midrule
    \textbf{Mean (16 configs)}        & \textbf{$+1.3$}  & \textbf{$+31.8$} & \textbf{$+25.1$} & \textbf{$+26.4$} \\
    \footnotesize 95\% bootstrap CI    & \scriptsize [$-4.6$, $+8.6$] & \scriptsize [$+28.8$, $+34.8$] & \scriptsize [$+20.8$, $+29.1$] & \scriptsize [$+22.5$, $+30.0$] \\
    \bottomrule
  \end{tabular}
\end{table}

\paragraph{Within-tier reading.} The $+32.6$~pp aggregate gap is
preserved at L2--L4 ($+31.8$, $+25.1$, $+26.4$~pp; every CI above
zero) and effectively vanishes only at L1, where the contextual pool
collapses to $n{=}13$ questions and ceiling effects dominate. The L4
result is the most informative: the contextual pool is large
($n{=}885$) and the gap is still $+26.4$~pp with a tight CI, so the
closed-book vs.\ source-grounded contrast is not an artefact of
difficulty composition --- it is a stable property of the slice
across non-trivial tiers, consistent with B2 identifying questions
recoverable from pre-training rather than questions that happen to be
easier-stated.

\FloatBarrier
\subsection{Per-domain breakdown}

\begin{table}[H]
  \centering
  \footnotesize
  \caption{Per-config per-domain accuracy (\%).}
  \label{tab:full-domain}
  \begin{tabular}{lrrrrrr}
    \toprule
    Config & Regions & Varieties & Producers & Viticulture & Winemaking & Business \\
    \midrule
    claude-opus-4.7              & 83.2 & 80.2 & 86.8 & 77.9 & 86.6 & 64.2 \\
    claude-haiku-4.5             & 55.8 & 49.7 & 58.7 & 54.2 & 56.7 & 37.8 \\
    gpt-5                        & 88.1 & 78.2 & 88.4 & 82.8 & 80.7 & 63.0 \\
    gpt-5-mini                   & 82.1 & 74.4 & 83.5 & 77.5 & 77.5 & 66.3 \\
    gemini-2.5-pro               & 87.0 & 78.8 & 89.0 & 78.5 & 77.5 & 61.8 \\
    gemini-2.5-flash             & 76.9 & 72.9 & 82.3 & 73.6 & 77.5 & 60.6 \\
    llama-3.3-70b                & 68.9 & 62.9 & 77.2 & 65.7 & 70.6 & 50.4 \\
    llama-3.1-8b                 & 62.2 & 55.6 & 69.5 & 61.7 & 59.4 & 47.6 \\
    deepseek-v3                  & 74.4 & 68.5 & 73.2 & 70.8 & 69.0 & 51.2 \\
    qwen-2.5-72b                 & 69.0 & 65.1 & 73.4 & 67.5 & 70.1 & 52.8 \\
    qwen-2.5-7b                  & 59.1 & 55.6 & 59.4 & 57.4 & 57.8 & 44.7 \\
    mistral-large-2411           & 71.5 & 65.6 & 75.4 & 69.4 & 73.3 & 51.6 \\
    o3                           & 88.4 & 81.3 & 90.9 & 80.3 & 76.5 & 65.4 \\
    gemini-2.5-pro-thinking      & 86.5 & 80.5 & 90.0 & 80.5 & 83.4 & 60.2 \\
    deepseek-r1                  & 80.5 & 73.6 & 85.6 & 72.2 & 76.5 & 64.6 \\
    claude-opus-4.7-thinking     & 83.2 & 80.9 & 87.0 & 78.7 & 84.0 & 61.4 \\
    \midrule
    \textbf{Mean (16 configs)}   & \textbf{76.0} & \textbf{70.2} & \textbf{79.4} & \textbf{71.8} & \textbf{73.6} & \textbf{56.5} \\
    \bottomrule
  \end{tabular}
\end{table}

\begin{figure}[H]
  \centering
  \includegraphics[width=0.65\linewidth]{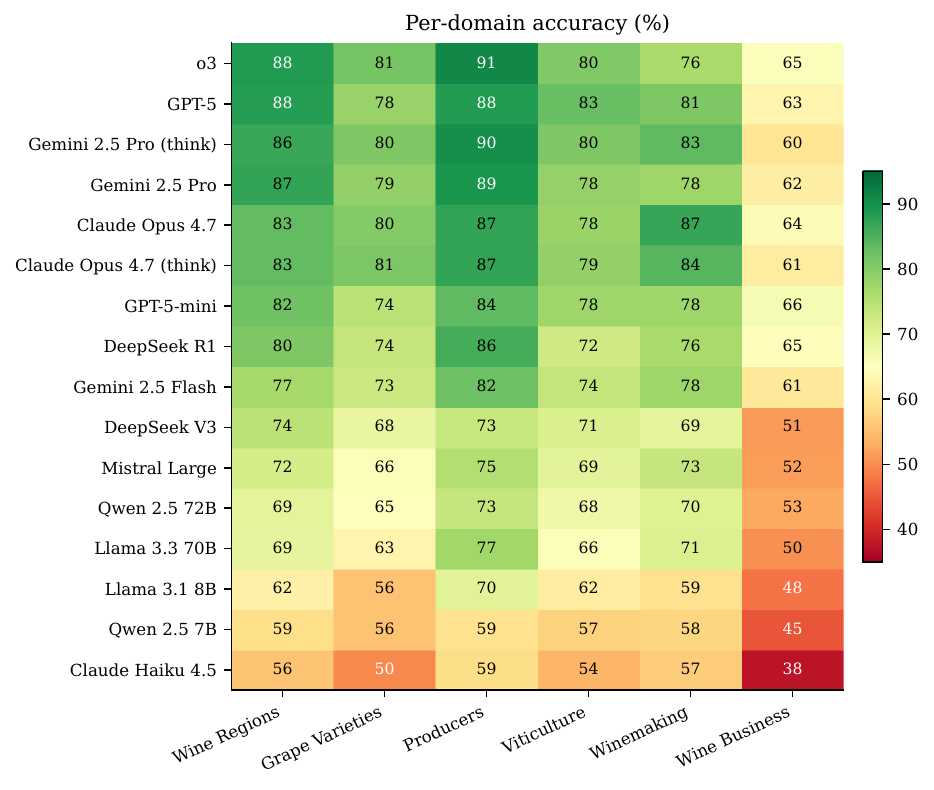}
  \caption{Per-domain accuracy as a heatmap. \emph{Wine business} is
    the hardest domain across every tier (mean 56.5\%) — both because
    business-domain facts are sparser in the training distribution and
    because the C4 difficulty re-label promoted business questions
    that are computationally harder (multi-fact reasoning over
    regulation).}
  \label{fig:domain-heatmap}
\end{figure}

\FloatBarrier
\subsection{Per-strategy breakdown}

\begin{table}[H]
  \centering
  \footnotesize
  \caption{Per-config per-strategy accuracy (\%).}
  \label{tab:full-strategy}
  \begin{tabular}{lrrrrr}
    \toprule
    Config & FTQ & Scenario & Template & Comparative & Distractor \\
    \midrule
    claude-opus-4.7              & 77.7 & 85.6 & 92.8 & 79.5 & 82.7 \\
    claude-haiku-4.5             & 47.5 & 62.7 & 70.7 & 53.3 & 56.8 \\
    gpt-5                        & 79.9 & 80.9 & 95.4 & 79.5 & 87.9 \\
    gpt-5-mini                   & 73.8 & 80.6 & 92.8 & 78.3 & 84.7 \\
    gemini-2.5-pro               & 77.4 & 87.5 & 93.1 & 82.4 & 86.4 \\
    gemini-2.5-flash             & 69.5 & 84.0 & 89.7 & 77.5 & 79.5 \\
    llama-3.3-70b                & 59.5 & 80.6 & 89.7 & 64.8 & 71.9 \\
    llama-3.1-8b                 & 52.2 & 70.5 & 89.7 & 60.2 & 63.7 \\
    deepseek-v3                  & 64.5 & 80.3 & 86.4 & 71.7 & 73.3 \\
    qwen-2.5-72b                 & 59.6 & 79.3 & 88.7 & 69.3 & 73.6 \\
    qwen-2.5-7b                  & 49.1 & 72.1 & 78.9 & 57.0 & 61.0 \\
    mistral-large-2411           & 62.7 & 80.6 & 87.4 & 67.6 & 73.3 \\
    o3                           & 79.9 & 84.3 & 94.9 & 84.8 & 88.4 \\
    gemini-2.5-pro-thinking      & 78.5 & 88.4 & 95.6 & 84.0 & 84.2 \\
    deepseek-r1                  & 73.4 & 75.5 & 92.5 & 77.0 & 80.7 \\
    claude-opus-4.7-thinking     & 77.3 & 85.6 & 93.6 & 80.3 & 83.0 \\
    \midrule
    \textbf{Mean (16 configs)}   & \textbf{67.6} & \textbf{79.9} & \textbf{89.5} & \textbf{73.0} & \textbf{76.9} \\
    \bottomrule
  \end{tabular}
\end{table}

\begin{figure}[H]
  \centering
  \includegraphics[width=0.6\linewidth]{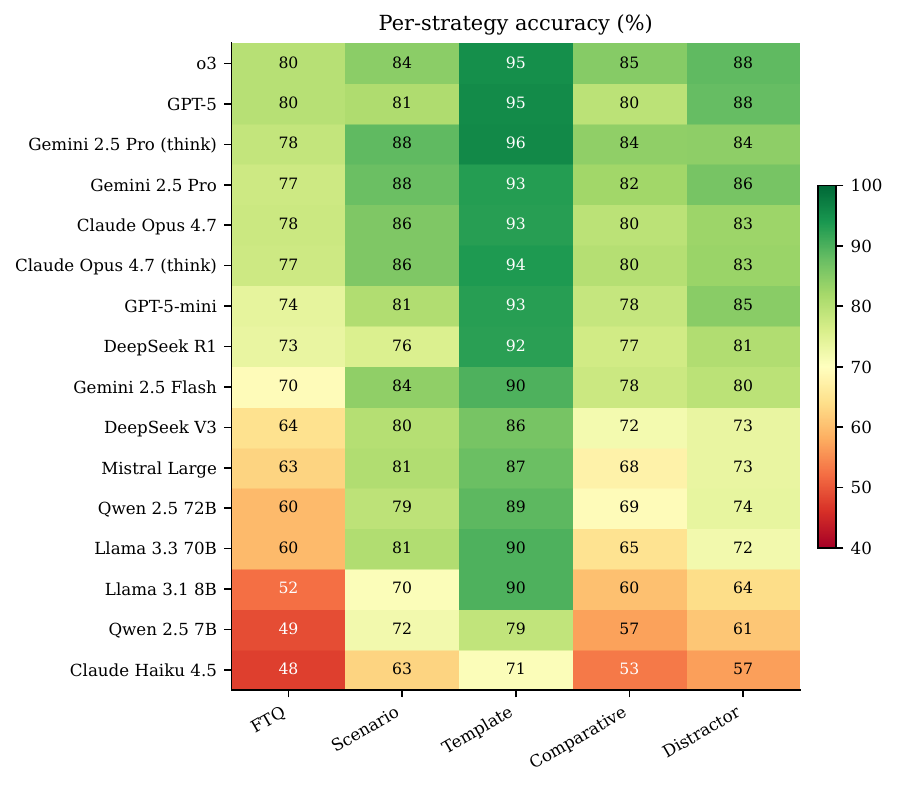}
  \caption{Per-strategy accuracy heatmap. Templates are the easiest
    strategy uniformly (89.5\% mean), even for small models, because
    they are deterministic single-fact recall questions. The
    fact-to-question strategy is hardest because its distractors are
    sampled from confusable same-category entities; comparative
    questions add the constraint that the model must integrate two
    facts.}
  \label{fig:strategy-heatmap}
\end{figure}

\FloatBarrier
\subsection{Per-difficulty breakdown}

\begin{table}[H]
  \centering
  \footnotesize
  \caption{Per-config per-difficulty accuracy (\%).}
  \label{tab:full-difficulty}
  \begin{tabular}{lrrrr}
    \toprule
    Config & L1 (entry) & L2 & L3 & L4 (expert) \\
    \midrule
    claude-opus-4.7              & 98.0 & 77.0 & 86.7 & 69.1 \\
    claude-haiku-4.5             & 74.3 & 48.3 & 59.9 & 38.8 \\
    gpt-5                        & 98.3 & 82.4 & 87.5 & 69.2 \\
    gpt-5-mini                   & 98.0 & 73.7 & 84.7 & 64.8 \\
    gemini-2.5-pro               & 98.6 & 79.2 & 87.3 & 68.5 \\
    gemini-2.5-flash             & 95.4 & 70.9 & 82.2 & 60.1 \\
    llama-3.3-70b                & 91.9 & 60.6 & 74.8 & 50.3 \\
    llama-3.1-8b                 & 84.8 & 55.4 & 65.3 & 45.0 \\
    deepseek-v3                  & 95.8 & 65.9 & 76.3 & 52.4 \\
    qwen-2.5-72b                 & 93.2 & 60.5 & 73.7 & 51.4 \\
    qwen-2.5-7b                  & 82.0 & 50.6 & 60.0 & 43.3 \\
    mistral-large-2411           & 94.4 & 63.1 & 73.3 & 54.0 \\
    o3                           & 98.4 & 81.4 & 89.4 & 71.2 \\
    gemini-2.5-pro-thinking      & 99.1 & 80.2 & 88.3 & 69.4 \\
    deepseek-r1                  & 97.5 & 74.2 & 82.0 & 62.1 \\
    claude-opus-4.7-thinking     & 98.3 & 76.8 & 87.0 & 68.6 \\
    \midrule
    \textbf{Mean (16 configs)}   & \textbf{93.6} & \textbf{68.8} & \textbf{78.7} & \textbf{58.7} \\
    \bottomrule
  \end{tabular}
\end{table}

\begin{figure}[H]
  \centering
  \includegraphics[width=0.55\linewidth]{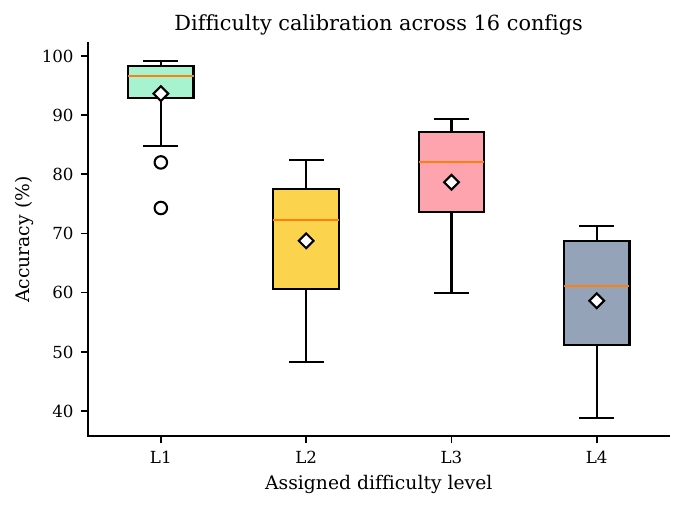}
  \caption{Difficulty calibration: distribution of accuracy across the
    16 configs at each level. The medians sit at L1 96\%, L2 70\%, L3
    81\%, L4 60\%.}
  \label{fig:difficulty-box}
\end{figure}

\vspace{-0.5em}
\paragraph{On the L2 / L3 / L4 means in Table~\ref{tab:full-difficulty}.}
Two features of the column averages deserve comment. First, the L2
mean (68.8\%) is \emph{below} the L3 mean (78.7\%) --- an apparent
inversion of the difficulty ordering. This is an artefact of the C4
difficulty re-label (Section~\ref{sec:audit}): many questions
originally labelled L2 by their generator were promoted to L3, leaving
behind a residual L2 set that is concentrated in the
\emph{wine\_business} pillar (the hardest domain by a wide margin,
Table~\ref{tab:full-domain}). The L3 set, in contrast, was enriched
with cleanly-stated multi-fact items from the higher-quality scrapers.
Second, the L3 $\rightarrow$ L4 drop is real and large
(78.7\%~$\rightarrow$~58.7\%, a 20~pp fall), reflecting that L4 is the
expert tier where the source fact is genuinely required to answer the
question; that fall is also where the inter-config spread is widest
(Section~\ref{sec:eval-rank}), making L4 the most discriminating slice
for distinguishing models. Both effects are intended outcomes of the
audit-driven difficulty re-label rather than miscalibration.

\section{Application Screenshots}
\label{app:screenshots}

\subsection{Human-review web application}

The human-review web application is the front-end through which
WSET-Diploma reviewers rate stratified gold-sheet questions and the
through which the agent-vs-human $\kappa$ values reported in
Section~\ref{sec:gold} are collected. It is a Flask application with
two layers of authentication (outer HTTP Basic for shared-link access,
inner per-reviewer session cookies) and writes ratings into a separate
\texttt{human\_reviews} Postgres table to keep gold-sheet labels
independent of automated audit findings. Each reviewer is associated
with a WSET certification level recorded at registration, and the app
supports multi-reviewer inter-rater agreement ($\kappa$) computation
out of the box.

\begin{figure}[H]
  \centering
  \includegraphics[width=0.62\linewidth]{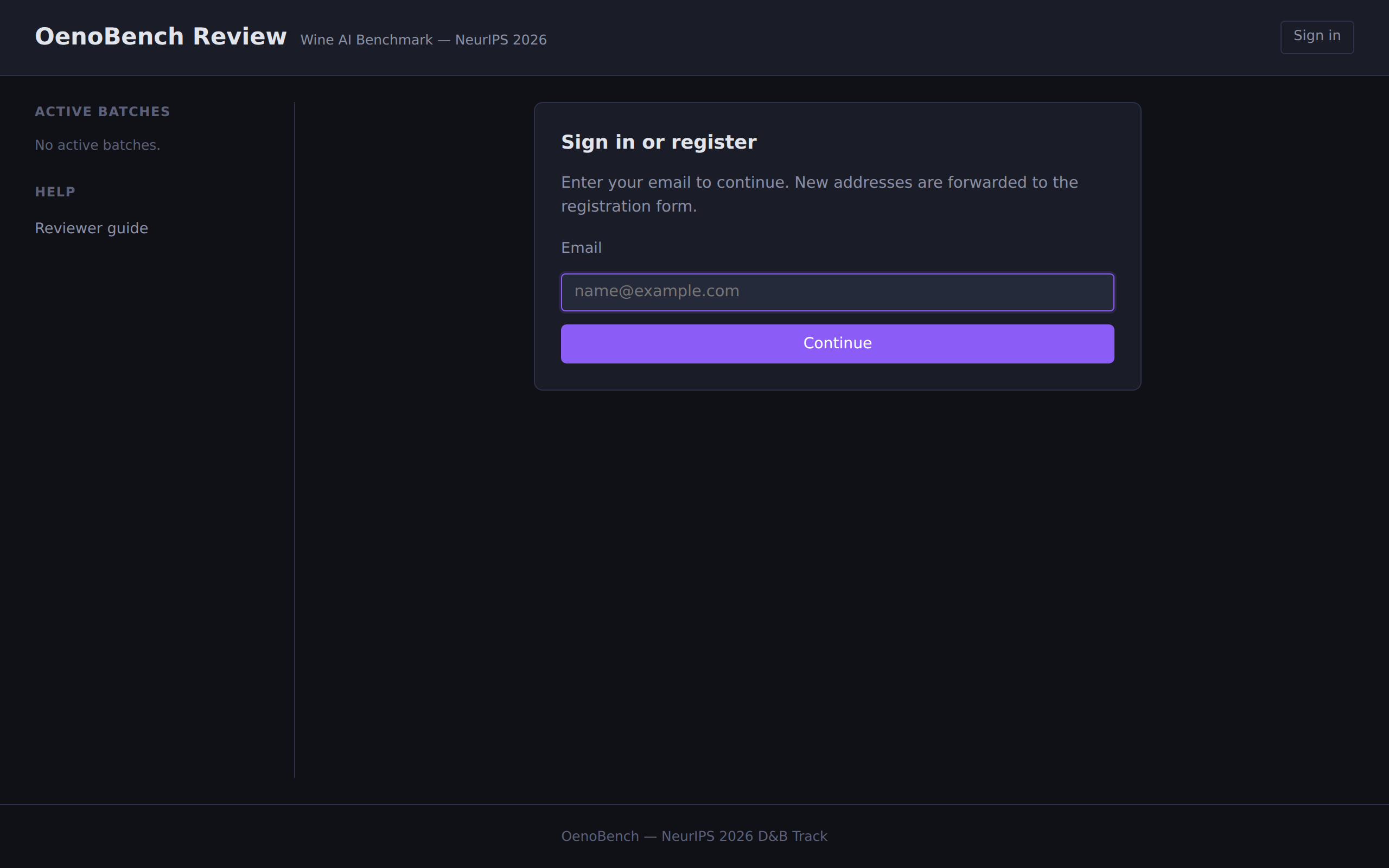}
  \caption{Sign-in / register screen. Outer HTTP Basic Auth gates
    shared-link access; the inner form establishes a per-reviewer
    session for IRR attribution.}
  \label{fig:review-login}
\end{figure}

\begin{figure}[H]
  \centering
  \includegraphics[width=0.62\linewidth]{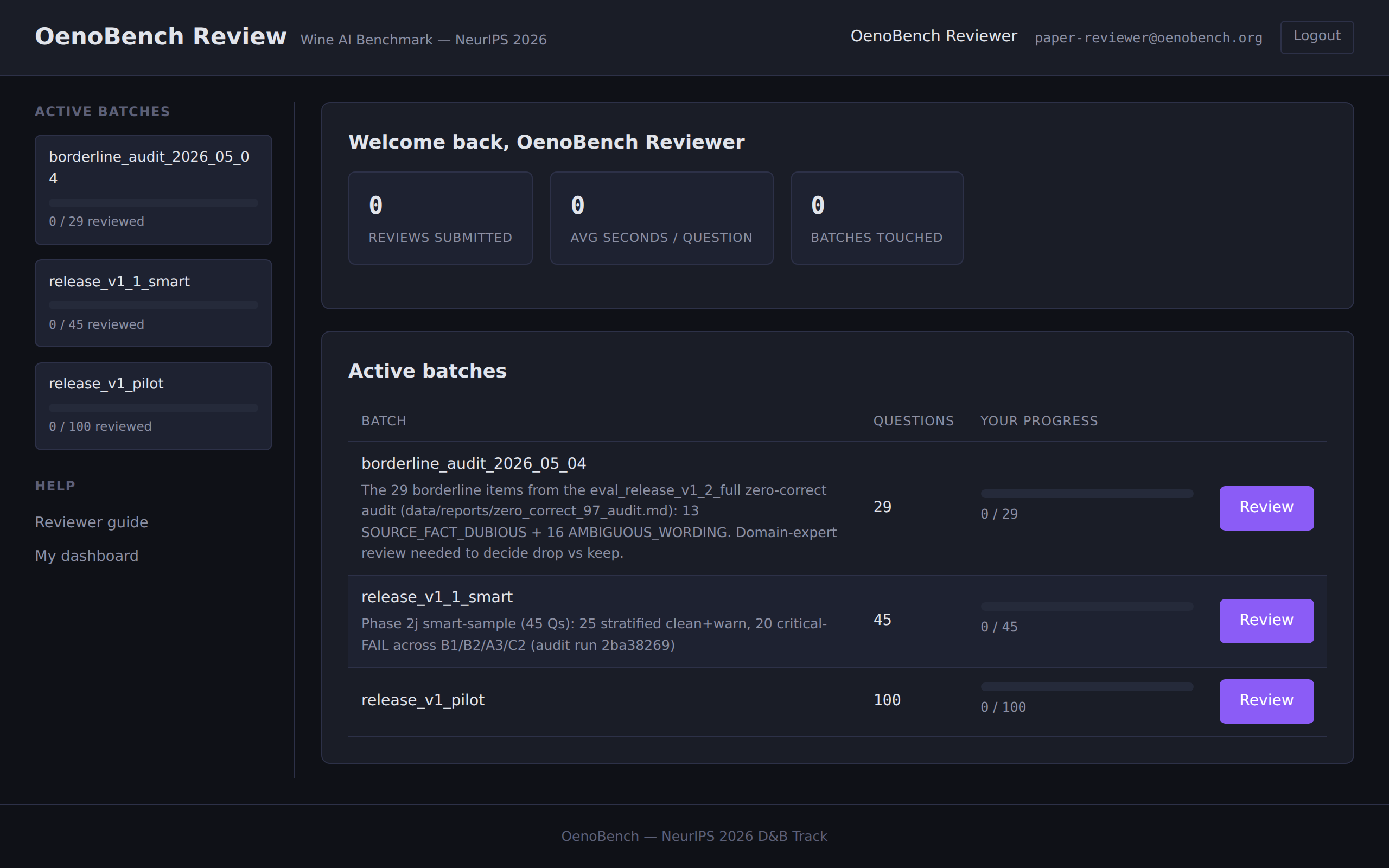}
  \caption{Reviewer dashboard. Active batches are listed; clicking a
    batch enters the rubric-scoring flow (eight-rubric grid with
    \textsc{pass}/\textsc{warn}/\textsc{fail} per rubric, inline
    reviewer-guide hints, and time-on-page tracking).}
  \label{fig:review-dashboard}
\end{figure}

\section{Released Artifacts and Reproducibility}
\label{app:artifacts}

\subsection{Released artifacts}

\begin{itemize}\itemsep -1pt
  \item \textbf{Dataset (HuggingFace).}
    \url{https://huggingface.co/datasets/oenobench/oenobench} —
    \texttt{release\_v1.2}, 3{,}266 questions, Parquet test split,
    \texttt{README.md} datasheet, \texttt{croissant.json} manifest.
    License: CC-BY-SA-4.0. Versioning is permanent: prior releases
    remain accessible via revision tags.
  \item \textbf{Code (GitHub).}
    \url{https://github.com/nikitahudov/oenobench} — full pipeline:
    35 scrapers, 5 question generators, 14 audit agents, the human
    review web app, and the monitoring dashboard. Apache 2.0.
  \item \textbf{Process log.} \texttt{docs/PROCESS\_LOG.md} —
    chronological lab-notebook of every phase shipped, with sources,
    methodology, quality-control counts, and decision rationale, in
    the format prescribed for the methodology sections of this paper.
  \item \textbf{Audit reports.} \texttt{docs/QUALITY\_AUDIT\_REPORT.md},
    \texttt{docs/RELEASE\_V1\_2\_AUDIT\_ACTIONS.md},
    \texttt{docs/GOLD\_CALIBRATION\_ANALYSIS.md},
    \texttt{data/reports/zero\_correct\_97\_audit.md}.
  \item \textbf{Eval reports.}
    \texttt{data/reports/eval\_release\_v1\_2\_full\_cleaned.md} (the
    16-config × 3{,}266-Q run reported in this paper).
\end{itemize}

\subsection{Reproducibility scripts}

The construction pipeline is driven by a small number of shell scripts
in \texttt{scripts/}:

\begin{itemize}\itemsep -1pt
  \item \texttt{run\_all\_scrapers.sh} — orchestrates the 35 scrapers
    end-to-end. Each scraper writes a timestamped log to
    \texttt{data/logs/}.
  \item \texttt{run\_release\_v1\_build.sh} — full generation pipeline
    invocation; idempotent and resume-safe via stamped tags.
  \item \texttt{export\_release\_v1\_2\_to\_parquet.py} — produces the
    HuggingFace Parquet directly from Postgres; the released file
    is the byte-identical output of this script.
  \item \texttt{build\_smart\_review\_sheet.py} — generates a 50-question
    stratified gold-sheet review batch (random + critical-FAIL +
    borderline-WARN strata).
  \item \texttt{tag\_audit\_actions.py} — applies the audit drop /
    relabel policy to a tagged corpus; this is the script that
    produced the \texttt{release\_v1.2} state from
    \texttt{release\_v1.1}.
\end{itemize}

\subsection{Database schema}

The released schema is initialised by the migrations under
\texttt{config/postgres/} (six idempotent files, applied in order on
container startup):
\texttt{init.sql} (sources, facts, questions, question\_facts,
generation\_metadata, evaluation\_runs, evaluation\_answers,
human\_reviews),
\texttt{002\_audit\_schema.sql} (audit\_runs, audit\_findings,
audit\_gold\_labels, audit\_severity enum),
\texttt{003\_sample\_schema.sql} (mirror schema for the Phase~5
sample-DB eval),
\texttt{003\_cb\_reserve.sql} (closed-book reserve pool tagging),
\texttt{004\_eval\_telemetry.sql} (provider, tokens, reasoning\_config,
latency columns on \texttt{evaluation\_answers}),
\texttt{005\_or\_cost\_telemetry.sql} (OpenRouter authoritative cost
columns).

The released schema is documented in code
(\texttt{config/postgres/}); the load-bearing tables are
\texttt{sources} (registry), \texttt{facts} (atomic facts grounded by
source), \texttt{questions} + \texttt{question\_facts} (corpus and
fact-question links), \texttt{generation\_metadata} (per-question
generator and strategy), \texttt{audit\_findings} (per-agent
verdicts), and \texttt{human\_reviews} (gold-sheet ratings).

\subsection{Compute and cost}

The full evaluation reported in Section~\ref{sec:eval} ran in
\textbf{120 minutes 34 seconds} on a single VM driving 16 OpenRouter
configurations in parallel. \textbf{Total LLM API spend across the
entire project was \$783}, paid out-of-pocket to OpenRouter and
covering all generation pilots, audit cycles, gold-sheet calibration
runs, and the final 16-config evaluation. No GPU compute was used;
all model inference was via OpenRouter (a unified API gateway). The
release artefacts include all per-call telemetry (provider, tokens,
reasoning configuration, p50/p95 latency, OR-authoritative cost) for
the full evaluation run; the per-config cost ledger is reproduced
inline as Table~\ref{tab:cost-ledger-app}.

\begin{table}[h]
  \centering
  \footnotesize
  \caption{Per-config cost ledger for the full 16-config evaluation
    (3{,}266 questions per config). Effective OpenRouter spend.}
  \label{tab:cost-ledger-app}
  \begin{tabular}{lrrr}
    \toprule
    Config & Total cost & Correct ans. & Cost / correct \\
    \midrule
    llama-3.1-8b               & \$0.01 & 1{,}976 & \$0.0000 \\
    llama-3.3-70b              & \$0.04 & 2{,}190 & \$0.0000 \\
    gemini-2.5-flash           & \$0.12 & 2{,}454 & \$0.0000 \\
    deepseek-v3                & \$0.12 & 2{,}295 & \$0.0001 \\
    qwen-2.5-7b                & \$0.12 & 1{,}860 & \$0.0001 \\
    qwen-2.5-72b               & \$0.15 & 2{,}202 & \$0.0001 \\
    claude-haiku-4.5           & \$0.44 & 1{,}741 & \$0.0003 \\
    mistral-large-2411         & \$0.87 & 2{,}256 & \$0.0004 \\
    gpt-5-mini                 & \$2.82 & 2{,}561 & \$0.0011 \\
    claude-opus-4.7            & \$3.35 & 2{,}647 & \$0.0013 \\
    claude-opus-4.7-thinking   & \$3.53 & 2{,}645 & \$0.0013 \\
    deepseek-r1                & \$10.54 & 2{,}517 & \$0.0042 \\
    o3                         & \$11.80 & 2{,}729 & \$0.0043 \\
    gemini-2.5-pro-thinking    & \$13.06 & 2{,}698 & \$0.0048 \\
    gpt-5                      & \$21.90 & 2{,}704 & \$0.0081 \\
    gemini-2.5-pro             & \$29.47 & 2{,}669 & \$0.0110 \\
    \bottomrule
  \end{tabular}
\end{table}

\subsection{Per-source licensing}
\label{sec:licensing}

Section~\ref{sec:sources} states that every fact in the corpus traces
to a public-record source or to a source distributed under a licence
that permits scraping and redistribution. We give the per-source
breakdown in Table~\ref{tab:licensing}, grouping the
35 scrapers' source families by tier and licence basis. Each row
maps to one or more rows in the released \texttt{sources} table, where
per-row \texttt{name}, \texttt{url}, \texttt{tier}, and access timestamp
are recorded for every fact (\texttt{facts.source\_id}~$\rightarrow$
\texttt{sources.id}). Wikipedia and Wikidata together account for 65.3\%
of facts and propagate their share-alike / public-domain dedications
through the \texttt{release\_v1.2} CC-BY-SA-4.0 licence. Government
records (Tier~1, 19.6\% of facts) are either explicit open-data
licences (INAO Licence Ouverte~2.0, EU re-use Decision 2011/833/EU) or
US-Government works in the public domain
(17~U.S.C.~\textsection 105: TTB, USDA Extension, UC IPM).

\begin{table}[h]
  \centering
  \footnotesize
  \caption{Per-source licensing for OenoBench's 35 scrapers, grouped by
    source family. ``Share'' is the percentage of the 38{,}104-fact
    corpus contributed; ``Tier'' is the source-of-authority label from
    Section~\ref{sec:sources}. Counts are computed directly from the
    \texttt{facts}~$\bowtie$~\texttt{sources} join in the released
    schema.}
  \label{tab:licensing}
  \begin{tabular}{p{0.31\linewidth}lp{0.30\linewidth}r}
    \toprule
    Source family & Tier & Licence / legal basis & Share \\
    \midrule
    Wikipedia (MediaWiki API)                                            & T2          & CC-BY-SA 4.0                                           & 34.4\% \\
    Wikidata (SPARQL)                                                    & T1 / T2     & CC0 1.0                                                & 30.3\% \\
    HuggingFace \texttt{spawn99/wine-reviews}                            & T2          & HuggingFace dataset terms (open redistribution)        &  8.5\% \\
    UC Davis (AVA Project, FPS, Wine Ontology)                           & T1          & Open / CC-BY (UC Davis Library)                        &  5.7\% \\
    UC IPM \& USDA Extension (eXtension, PSU/OSU)                        & T1 / T2     & Public domain --- US Gov.\ work (17 USC \S105)         &  4.7\% \\
    Kaggle: wine-reviews (Wine Enthusiast scrape)                        & T3          & Kaggle dataset terms (research use)                    &  3.8\% \\
    INAO (data.gouv.fr open-data)                                        & T1          & Licence Ouverte 2.0 (Etalab)                           &  3.9\% \\
    Wine consortia + national bodies (CIVB, BIVB, CIVC, IT consortia, NZ Wine, Austrian Wine, Deutsches Weininstitut, Spanish DO) & T2 & Industry-body publications, educational use & 3.2\% \\
    OENO One                                                             & T2          & CC-BY 4.0 (open-access journal)                        &  2.0\% \\
    TTB (27 CFR Part 4, Established AVAs, Approved Grape Names)          & T1          & Public domain --- US Gov.\ work                        &  1.3\% \\
    National ministries + EU registries (MASAF, MAPA, IVV, IVDP, EUR-Lex, GIView) & T1 / T2 & Government public records / EU Decision 2011/833/EU & 1.3\% \\
    OIV publications (Code of Oenological Practices, statistics)         & T1          & Open access --- intergov.\ organisation                &  0.2\% \\
    Vitis (Hochschule Geisenheim)                                        & T2          & Open access                                            &  0.3\% \\
    National reference databases (CA, UK, HR/SI, HU/GE, LB/IL, +regions) & T2          & Reference / official-body content                      &  0.3\% \\
    Kaggle: wine-quality (UCI ML Repository)                             & T2          & CC-BY 4.0 (UCI)                                        &  0.2\% \\
    \bottomrule
  \end{tabular}
\end{table}

The released corpus, including this paper's quoted statistics, is
distributed under \textbf{CC-BY-SA 4.0}. The construction code is
released under \textbf{Apache 2.0}. Verbatim source text is never
stored: every fact is paraphrased atomically (Section~\ref{sec:extraction})
and a paraphrase guard at audit time (agent~A3) re-checks for
verbatim leakage. We retain \texttt{accessed\_date} and a per-source
\texttt{content\_date} for every fact, so dataset users can audit
freshness against original sources.

\subsection{Test status}

771 of 771 unit tests pass on \texttt{main} as of the camera-ready
build. Tests cover the fact-processing pipeline, the five generation
strategies, all 14 audit agents, the orchestrator, the closed-book
gate, the eval harness (slot dispatch, cost computation,
reasoning-mode prompt assembly), and the review-app endpoints.

\section{Construction Timeline}
\label{app:timeline}

The full chronological lab notebook of construction is in
\texttt{docs/PROCESS\_LOG.md}. We summarise the major dated milestones
here for paper-traceable provenance.

\begin{table}[h]
  \centering
  \footnotesize
  \caption{Construction-and-evaluation timeline (selected milestones).}
  \label{tab:timeline}
  \begin{tabular}{llp{0.7\linewidth}}
    \toprule
    Phase & Dates & Milestone \\
    \midrule
    0   & 2026-02-13         & Repository initialised: Docker stack
                               (Postgres / Elasticsearch / Neo4j / Redis), schema, and the
                               first four scrapers (Wikidata, HuggingFace, Kaggle,
                               UC Davis) committed. \\
    1   & 2026-02-13 -- 04-01 & 35 scrapers built incrementally
                               (INAO, TTB, EU/OIV, Wikipedia, Italian / Spanish /
                               Portuguese / German / Austrian / Greek registries,
                               Bordeaux / Burgundy / Champagne / Italian
                               consortia, USDA Extension, OENO One / Vitis /
                               AJEV academic crawlers, country enrichment passes);
                               first $\sim$16{,}702-fact working corpus
                               assembled by end of March. \\
    1.1 & 2026-04-07         & \emph{Provenance audit reveals 19 scrapers contained
                               hardcoded LLM-generated facts}; corpus paused. \\
    1.2 & 2026-04-08 -- 04-11 & Shared infrastructure
                               (\texttt{\_fact\_processing.py},
                               \texttt{\_web\_helpers.py},
                               \texttt{\_wiki\_helpers.py}) shipped;
                               Phase~1 rebuild fixes 8 quality-issue scrapers; Phase~2
                               rebuild replaces all 17 hardcoded scrapers with genuine
                               Wikipedia + SPARQL + official sources. \emph{38{,}104
                               fact corpus restored, all provenance-traceable.} \\
    2   & 2026-04-12 -- 04-22 & Five generation strategies built and iteratively
                               tuned; multi-model orchestrator with quota management;
                               closed-book gate v2 wired into all five generators. \\
    2c  & 2026-04-18 -- 05-02 & Nine-agent audit framework shipped across 4 teams;
                               sixteen audit pilots (v1--v16) tune thresholds and
                               speedups; 472/472 → 771/771 tests pass. \\
    2g  & 2026-04-28 -- 05-02 & Ten generation-pipeline speedup levers shipped
                               (LLM cache, in-process dispatch, ThreadPool dispatch,
                               substantiveness filter, circuit breaker, tier-aware
                               gate). v9 build wall: 18~min vs.\ v8's 11.4~h. \\
    2j  & 2026-05-03         & \texttt{release\_v1.1} (3{,}670 candidate questions)
                               assembled from staged generation runs and the curated
                               sample-DB. Audit cycle drops 341 questions on critical
                               FAILs and re-labels difficulty for 1{,}259 (51\%
                               L3+L4 vs.\ prior 14\%). Cycle cost \$76 / 5h~22m. \\
    4   & 2026-05-03         & Human-review web app shipped on port 5556 with
                               separate \texttt{human\_reviews} schema and
                               multi-reviewer IRR support. \\
    5   & 2026-05-03 -- 05-04 & Full evaluation: 16 configs $\times$ 3{,}329 questions
                               in 120m~34s for \$98.33; post-eval audit of 97
                               zero-correct items removes 54 defects + 9
                               borderline-review drops; \texttt{release\_v1.2}
                               locked at 3{,}266 questions. \\
    HF  & 2026-05-04         & Dataset card + Croissant manifest + Parquet test
                               split published; this paper drafted. \\
    \bottomrule
  \end{tabular}
\end{table}

\newpage
\section*{NeurIPS Paper Checklist}

\begin{enumerate}

\item {\bf Claims}
    \item[] Question: Do the main claims made in the abstract and introduction accurately reflect the paper's contributions and scope?
    \item[] Answer: \answerYes{}
    \item[] Justification: Section~\ref{sec:intro} lists four contributions, and each is supported by a numbered section: dataset construction (Section~\ref{sec:dataset}), automated audit (Section~\ref{sec:audit}), bias-aware evaluation (Section~\ref{sec:eval}). The abstract's headline numbers (3{,}266 questions, 16-config eval, +32.6~pp closed-book vs.\ contextual gap) are reproduced with full tables in Appendix~\ref{app:eval}.

\item {\bf Limitations}
    \item[] Question: Does the paper discuss the limitations of the work performed by the authors?
    \item[] Answer: \answerYes{}
    \item[] Justification: Section~\ref{sec:limitations} discusses three limitations explicitly: snapshot vs.\ moving target, B2 audit calibration ceiling, and residual self-preference.

\item {\bf Theory assumptions and proofs}
    \item[] Question: For each theoretical result, does the paper provide the full set of assumptions and a complete (and correct) proof?
    \item[] Answer: \answerNA{}
    \item[] Justification: This is a dataset-and-benchmark paper; no formal theoretical results are claimed.

\item {\bf Experimental result reproducibility}
    \item[] Question: Does the paper fully disclose all the information needed to reproduce the main experimental results of the paper?
    \item[] Answer: \answerYes{}
    \item[] Justification: Appendix~\ref{app:artifacts} lists the construction scripts, the dataset URL, the model slate, the per-call telemetry available alongside the released corpus (including run id, wall time, OR-authoritative cost, and exact config strings), and the deterministic build pipeline. Section~\ref{sec:eval} specifies the 16-config slate, the single-letter (A--D) output protocol, and the bootstrap procedure.

\item {\bf Open access to data and code}
    \item[] Question: Does the paper provide open access to the data and code, with sufficient instructions to faithfully reproduce the main experimental results, as described in supplemental material?
    \item[] Answer: \answerYes{}
    \item[] Justification: Code at \url{https://github.com/nikitahudov/oenobench} (Apache~2.0); data at \url{https://huggingface.co/datasets/oenobench/oenobench} (CC-BY-SA-4.0). Exact reproduction commands are documented in Appendix~\ref{app:artifacts}.

\item {\bf Experimental setting/details}
    \item[] Question: Does the paper specify all the training and test details necessary to understand the results?
    \item[] Answer: \answerYes{}
    \item[] Justification: Section~\ref{sec:eval} specifies the 16-config slate, single-letter output protocol (A--D, \texttt{max\_tokens=5}, five-stop fallback). Appendix~\ref{app:eval} gives full per-config × per-domain × per-strategy × per-difficulty tables. Appendix~\ref{app:methodology} gives audit-agent thresholds and prompt summaries.

\item {\bf Experiment statistical significance}
    \item[] Question: Does the paper report error bars suitably and correctly defined or other appropriate information about the statistical significance of the experiments?
    \item[] Answer: \answerYes{}
    \item[] Justification: Reasoning lift, self-preference scores, and closed-book vs.\ contextual deltas are all reported with 95\% bootstrap confidence intervals (1{,}000 resamples). Section~\ref{sec:eval} states the procedure; Appendix~\ref{app:eval} (Tables~\ref{tab:reasoning-by-diff}, \ref{tab:sps-full}, \ref{tab:sps-by-tier}, \ref{tab:cb-by-tier}) reports per-config CIs.

\item {\bf Experiments compute resources}
    \item[] Question: For each experiment, does the paper provide sufficient information on the computer resources needed to reproduce the experiments?
    \item[] Answer: \answerYes{}
    \item[] Justification: Appendix~\ref{app:artifacts} reports the full 16-config evaluation wall time (120~min~34~s), total LLM calls (52{,}256), and effective evaluation cost (\$98.33), and the deployment topology (single VM driving 16 OpenRouter configurations in parallel; no GPU compute). The total OpenRouter spend across the entire project --- generation pilots, audit cycles, gold-sheet calibration, and the final evaluation --- was \textbf{\$783}, paid out-of-pocket and itemised in Appendix~\ref{app:artifacts} and the Datasheet (Appendix~\ref{app:datasheet}, Motivation).

\item {\bf Code of ethics}
    \item[] Question: Does the research conducted in the paper conform, in every respect, with the NeurIPS Code of Ethics?
    \item[] Answer: \answerYes{}
    \item[] Justification: All sources used are public-record; no human subjects research was conducted (gold-sheet ratings were produced by three WSET-certified reviewers including the lead author, treated as data-quality assurance); responsible-use guidance is given in the Datasheet (Appendix~\ref{app:datasheet}, ``Uses'') and in Section~\ref{sec:limitations}.

\item {\bf Broader impacts}
    \item[] Question: Does the paper discuss both potential positive societal impacts and negative societal impacts of the work performed?
    \item[] Answer: \answerYes{}
    \item[] Justification: The Datasheet (Appendix~\ref{app:datasheet}) covers alcohol-context and discouraged uses; source-licensing and the non-fabrication guarantee are documented in Section~\ref{sec:sources} and Table~\ref{tab:licensing} (Appendix~\ref{app:artifacts}); the corpus contains only public-record information about commercial wine entities (no private individuals; Datasheet ``Collection process''). Section~\ref{sec:future} discusses positive applications including specialist-model fine-tuning and pipeline retargeting to other expert domains.

\item {\bf Safeguards}
    \item[] Question: Does the paper describe safeguards that have been put in place for responsible release of data or models that have a high risk for misuse?
    \item[] Answer: \answerYes{}
    \item[] Justification: The dataset is text-only multiple-choice questions with public-record entity names; no scraped images or personal-data risks. We document discouraged uses in the Datasheet (Appendix~\ref{app:datasheet}, ``Uses'') including the recommendation against retargeting the pipeline to high-stakes domains without expert review.

\item {\bf Licenses for existing assets}
    \item[] Question: Are the creators or original owners of assets used in the paper properly credited and are the license and terms of use explicitly mentioned and properly respected?
    \item[] Answer: \answerYes{}
    \item[] Justification: The full per-source licensing table (Table~\ref{tab:licensing} in Appendix~\ref{app:artifacts}) credits every source family with licence and legal basis (Wikipedia CC-BY-SA 4.0, Wikidata CC0, INAO Licence Ouverte 2.0, TTB / USDA / UC IPM public domain (US Government work, 17~USC~\textsection 105), EU re-use Decision 2011/833/EU, OENO One CC-BY 4.0, etc.) and is also summarised in Section~\ref{sec:sources}. Per-fact source URL, accessed-date, and content-date are released alongside the corpus.

\item {\bf New assets}
    \item[] Question: Are new assets introduced in the paper well documented and is the documentation provided alongside the assets?
    \item[] Answer: \answerYes{}
    \item[] Justification: A full Datasheet is in Appendix~\ref{app:datasheet}; a Croissant manifest \citep{koehn2024croissant} ships with the HuggingFace release; all 14 audit agents (9 always-run, 5 escalation-gated; Section~\ref{sec:audit-arch}), 5 generation strategies, 5 generator-model families, and 35 scrapers are documented in source.

\item {\bf Crowdsourcing and research with human subjects}
    \item[] Question: For crowdsourcing experiments and research with human subjects, does the paper include the full text of instructions given to participants and screenshots, if applicable, as well as details about compensation?
    \item[] Answer: \answerNA{}
    \item[] Justification: No crowdsourcing or third-party human-subjects research. Gold-sheet ratings were produced by three WSET-certified reviewers (Diploma, Level~3, Level~2) including the lead author; the review web app screenshots in Appendix~\ref{app:screenshots} document the rubric and instructions.

\item {\bf Institutional review board (IRB) approvals or equivalent for research with human subjects}
    \item[] Question: Does the paper describe potential risks incurred by study participants, whether such risks were disclosed to the subjects, and whether IRB approvals were obtained?
    \item[] Answer: \answerNA{}
    \item[] Justification: As noted in Appendix~\ref{app:datasheet}, the work involves no human subjects research distinct from data-quality assurance by three WSET-certified reviewers including the lead author.

\item {\bf Declaration of LLM usage}
    \item[] Question: Does the paper describe the usage of LLMs if it is an important, original, or non-standard component of the core methods in this research?
    \item[] Answer: \answerYes{}
    \item[] Justification: LLMs are central to the methodology (multi-strategy question generation across five model families, nine-agent automated audit including a tri-judge panel, closed-book solvability pre-screen). Section~\ref{sec:dataset} and Section~\ref{sec:audit} describe their usage; Section~\ref{sec:sources} and Appendix~\ref{app:methodology} document the explicit fact-grounding constraint that prevents LLMs from being used as the source of factual claims.

\end{enumerate}

\end{document}